\documentclass{article}

\usepackage[pagebackref=true,breaklinks=true,colorlinks,urlcolor=blue,citecolor=blue,linkcolor=blue,bookmarks=false]{hyperref}
\usepackage{natbib}
\setcitestyle{numbers,square}
\usepackage[preprint]{neurips_data_2023}

\usepackage[utf8]{inputenc} 
\usepackage[T1]{fontenc}    
\usepackage{url}            
\usepackage{booktabs}       
\usepackage{amsfonts}       
\usepackage{nicefrac}       
\usepackage{microtype}      
\usepackage{xcolor}         
\usepackage{xspace}
\usepackage{amsmath}
\usepackage{bbm}
\usepackage{hyperref}
\usepackage{array}          

\usepackage{subcaption}
\usepackage[font=small]{caption}

\usepackage{float}
\usepackage{lscape}

\usepackage[lined,ruled,linesnumbered]{algorithm2e}

\usepackage{booktabs}
\usepackage{multirow}

\usepackage{paralist}
\usepackage{enumitem}

\usepackage{bm}
\usepackage{epsfig}
\usepackage{mathtools}
\usepackage{amsmath}
\usepackage{amssymb}
\usepackage{amsfonts}

\usepackage{color,colortbl}
\usepackage{siunitx} 

\usepackage{comment}

\usepackage{url}  

\usepackage{listings}

\usepackage{xspace}
\usepackage{setspace}

\usepackage{nicefrac}
\usepackage{microtype}
\usepackage[utf8]{inputenc} 
\usepackage[T1]{fontenc}    

\def\etal{\emph{et~al}\mbox{.}\ }
\def\ie{i.e.,\ }               
\def\aka{aka,\ }

\makeatletter 
  \newcommand\figcaption{\def\@captype{figure}\caption} 
  \newcommand\tabcaption{\def\@captype{table}\caption} 
\makeatother

\definecolor{SceneBlue}{rgb}{0.7031,    0.7812,    0.8632}
\definecolor{SceneRed}{rgb}{0.8867,    0.6171,    0.5781}
\definecolor{SceneOrange}{rgb}{0.9296,    0.8125,    0.5859}
\definecolor{ScenePurple}{rgb}{0.7265,    0.6289,    0.7773}
\definecolor{SceneGreen}{rgb}{0.6562,    0.7656,    0.5937}

\newcommand{\heading}[1]{\noindent\textbf{#1}}

\usepackage{wrapfig}

\usepackage{layouts}
\makeatletter 
\newcommand{\layoutdetails}{%
\begin{tabular}{ll}
 \texttt{\textbackslash{textwidth}} & \printinunitsof{in}\prntlen{\textwidth} \\
\texttt{\textbackslash{linewidth}} & \printinunitsof{in}\prntlen{\linewidth} \\
Main text font &  \f@size pt \f@family \\
\sffamily \small Caption text font &  \sffamily \small \f@size pt \f@family \\
\end{tabular}%
}

\newlength\paramargin
\newlength\figmargin
\newlength\tablemargin
\newlength\secmargin
\newlength\figcapmargin
\newlength\rowmargin
\newlength\tablecapmargin

\newcolumntype{a}{>{\columncolor[HTML]{EFEFEF}}c}
\newcolumntype{b}{>{\columncolor{yellow}}c}

\title{The MI-Motion Dataset and Benchmark for 3D Multi-Person Motion Prediction}

\author{%
  Xiaogang Peng$^{1}$,
  Xiao Zhou$^1$,
  Yikai Luo$^1$,
  Hao Wen$^2$,
  Yu Ding$^3$,
  Zizhao Wu$^{1}$\thanks{Corresponding author}~~
  \AND
  \vspace{-8 mm}
  \\
  \textsuperscript{1}Hangzhou Dianzi University
  \textsuperscript{2}National University of Defense Technology\\
  \textsuperscript{3}Netease Fuxi AI Lab\\
  \vspace{-25 mm}
}

\begin{document}

\maketitle

\begin{figure}[h]
  \centering
  \vspace{-1.0cm}
  \includegraphics[width=0.9\linewidth]{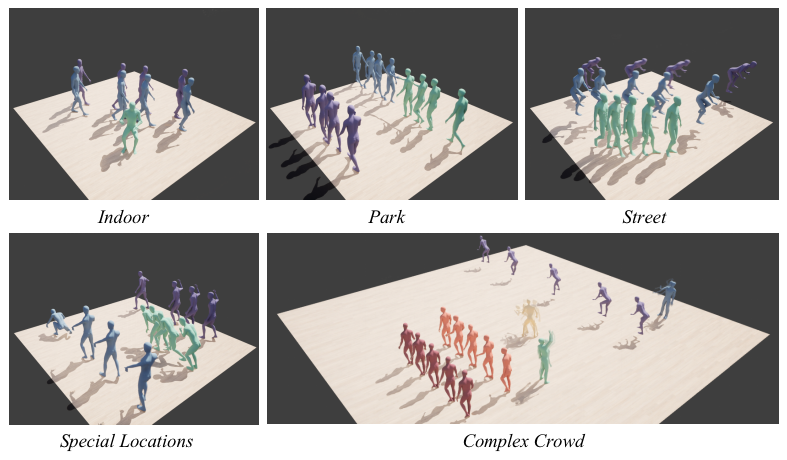}
  \caption{
  \textbf{The MI-Motion dataset} is categorized into 5 different daily activity scenes as shown here, and contains over 167k sequence frames with 3$\sim$6 interactive subjects. This dataset is articulately designed and constructed for 3D multi-person motion prediction.
  }
  \label{fig:teaser}
\end{figure}


\begin{abstract}

3D multi-person motion prediction is a challenging task that involves modeling individual behaviors and interactions between people. Despite the emergence of approaches for this task, comparing them is difficult due to the lack of standardized training settings and benchmark datasets. In this paper, we introduce the Multi-Person Interaction Motion (MI-Motion) Dataset, which includes skeleton sequences of multiple individuals collected by motion capture systems and refined and synthesized using a game engine. The dataset contains 167k frames of interacting people's skeleton poses and is categorized into 5 different activity scenes. To facilitate research in multi-person motion prediction, we also provide benchmarks to evaluate the performance of prediction methods in three settings: short-term, long-term, and ultra-long-term prediction. Additionally, we introduce a novel baseline approach that leverages graph and temporal convolutional networks, which has demonstrated competitive results in multi-person motion prediction. We believe that the proposed MI-Motion benchmark dataset and baseline will facilitate future research in this area, ultimately leading to better understanding and modeling of multi-person interactions.
\end{abstract}

\begin{figure}[h]
  \centering
  \includegraphics[width=0.9\linewidth]{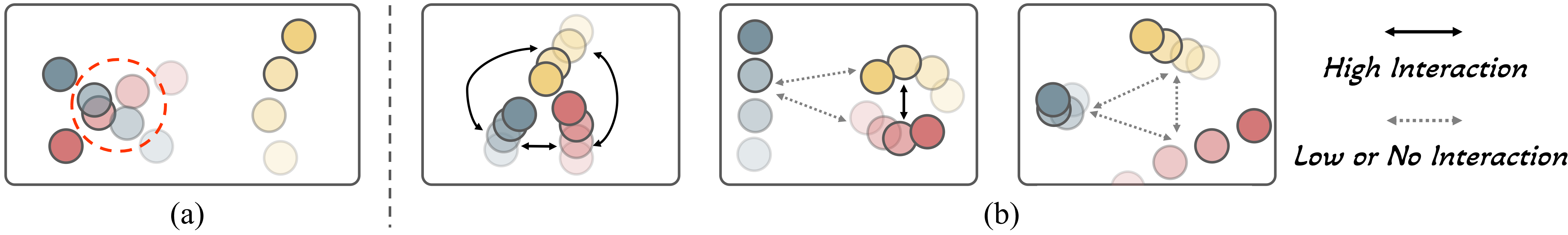}
  \caption{(a) Generating multi-person motion data by random mixing can sometimes result in motion collisions, which compromise the accuracy and authenticity of the data samples. (b) In MI-Motion dataset, each activity scene contains interactive motion samples with three different interaction levels: high-high interaction (left figure), low-high interaction (middle figure) and low-low interaction (right figure).}
  \label{fig:interact_level}
\end{figure}

\vspace{-2mm}
\section{Introduction}
Understanding and forecasting human motion is of significant importance in the fields of artificial intelligence and computer vision, with applications ranging from robot planning and autonomous driving to video surveillance \cite{gui2018adversarial, barsoum2018hp, li2020dynamic, martinez2017human, kundu2019bihmp, cao2020long}. In recent years, there has been a growing interest in the research area of multi-person motion prediction \cite{guo2022multi, wang2021multi, peng2023trajectory, adeli2020socially, adeli2021tripod, vendrow2022somoformer}. This task involves predicting the future poses of multiple individuals based on the past observations, taking into account their interactions and interdependencies. Unlike traditional approaches that focus on modeling individuals in isolation, multi-person motion prediction emphasizes the need to capture the complex interactions between subjects. 

While substantial progress has been made in single-person motion prediction, with the availability of large-scale benchmark datasets \cite{ionescu2013human3, CMU-Mocap} and advanced algorithms \cite{aksan2021spatio, mao2020history, Hernandez_Gall_Moreno_2020, gui2018adversarial, li2020dynamic, martinez2017human, kundu2019bihmp}, the field of multi-person motion prediction still faces challenges. One major hurdle is the lack of a comprehensive benchmark dataset specifically designed for multi-person motion prediction, especially considering the unique complexities introduced by the interactions between multiple subjects. Retrospectively, the task of multi-person motion prediction was initially introduced by Wang \etal \cite{wang2021multi}. They utilized CMU-Mocap dataset \cite{CMU-Mocap} to construct a mixed dataset that randomly combines single subject and two interactive subjects into a 3D scene. However, the limited availability of interactive samples in the CMU-Mocap \cite{CMU-Mocap} resulted in the mixed dataset that is relatively small in scale and lacks natural and diverse samples. Subsequently, Guo \etal \cite{guo2022multi} introduce the Extreme Pose Interaction (ExPI) dataset, which consists of 30K pose frames capturing the interactions between two professional dancers performing Lindy-hop dancing. However, it is worth noting that the dataset's action types and data scale are still limited in scope. 

In this paper, we present the MI-Motion (Multi-person Interaction Motion) dataset, a large dataset of multiple (3$\sim6$) subjects performing different interactions in 5 different daily activity scenes. To enable these interactive actions, we collected interactive action packs from the Unreal Engine asset store and capture certain specialized actions using a marker-based motion capture system. Instead of randomly mixing actions like \cite{wang2021multi}, which can result in unfeasible data as shown in Figure~\ref{fig:interact_level}-(a), we adopt a careful approach to customize interactive motion using these action sequences within the Unreal Engine 5 game engine\footnote{https://www.unrealengine.com/en-US/}. Futhermore, we refine and enhance the motion using the animation editor to create natural interactive behaviors. This meticulous process ensures that the generated motions are realistic and plausible, providing a more accurate representation of interactive scenarios. The whole synthesized dataset comprises 210 sequences featuring 3 to 6 subjects categorized by 5 different activity scenes (\textit{Park, Street, Indoor, Special Locations and Complex Crowd}), as illustrated in Figure 1. The 3D keypoints of interacting subjects are recorded across a total of 167K frames. To ensure a robust evaluation, we establish appropriate train/test splits and introduce five evaluation metrics for the multi-person motion prediction task. 


The MI-Motion dataset serves as a suitable resource for investigating the multi-person motion prediction task. To evaluate its effectiveness, we conduct a comprehensive comparison between MI-Motion and several existing motion datasets. We employ three state-of-the-art code-released deep learning models and proposed a novel convolution-based method termed Social Temporal Graph Convolutional Network (SocialTGCN). Then, we benchmark these models using the MI-Motion dataset under three prediction scenarios: short-term, long-term, and ultra-long-term prediction. Through extensive experiments and a thorough comparison against the MI-Motion dataset, we demonstrate that multi-person motion prediction remains a challenging and open problem, offering ample opportunities for future advancements and contributions.

\heading{The main contribution of the paper is twofold:} (1) We present a motion dataset of multiple interactive subjects with different interaction levels for the multi-person motion prediction task. Our dataset is publicly available at \href{https://mi-motion.github.io/}{\texttt{https://mi-motion.github.io/}}. (2) On the MI-Motion dataset, we benchmark three recent-released state-of-the-art methods and a novel baseline we proposed under various prediction settings, and provide accompanying code to ensure reproducibility and facilitate future research in this field.

\begin{table}[t] 
  \caption{\textbf{Overview of the publicly available datasets on 3D human motion prediction.} PE and MP indicate pose estimation and motion prediction, respectively. CMU-Mocap* denotes its category of human interaction, which only contains two interacting subjects. RGB-based datasets for pose estimation often exhibit significant noise, such as pose jittering.}
  \label{tab:dataset}
  \renewcommand\arraystretch{1.4}
  \centering
  \resizebox{\linewidth}{!}{
  \begin{tabular}{lccccccc}
    \toprule
     DataSet & Frames  &Sequence  &KeyPoints  &Person Count      &Human Pose from   &Task &Year \\
    \midrule
    H3.6M\cite{ionescu2013human3}    & 3600k      &-        &32        &1           &Marker-based  &PE \& MP            &’16   \\
    
    CMU-Mocap\cite{CMU-Mocap} & $\approx$600k      &-      &31       &1             &Marker-based    & MP  &’16    \\
    CMU-Mocap*\cite{CMU-Mocap} & $\approx$3k      &-      &31       &2              &Marker-based    & MP  &’16    \\
    
    MuPots-3D \cite{Mehta2017SingleShotM3} &8k        &20       &14       &2$\sim$3         &RGB    & PE \& MP         &’17 \\
    
    3DPW \cite{vonMarcard2018}     & 51k      &60      &17        &2            &RGB+IMU  & PE \& MP                &’18    \\

    AMASS \cite{Mahmood2019AMASSAO}    & -        & 11265      &37$\sim$91       &1           &Marker-based    & PE \& MP       &’19 \\
    
    GTA-IM \cite{caoHMP2020}  &1000k      &-       &98        &1            &Game engine      & MP    &'20   \\  
    
    ExPI \cite{guo2022multi}      & 30k      &115      &18       &2             &RGB   & MP              &’21    \\

     
     GIMO \cite{Zheng2022GIMOGH}   & 129k      &217      &53       &1             &IMU    & MP          &’22 \\
        
     \midrule
    MI-Motion (ours)     & 167k     &210      &20      &3 $\sim$ 6         &Game engine          & MP       &'23 \\
    \bottomrule
  \end{tabular}}
\end{table}

\section{Related Work}
\subsection{Datasets for 3D Human Motion Prediction}
\textbf{Single-person based datasets.} There are four widely-used motion datasets for single motion prediction: Human3.6M \cite{ionescu2013human3}, CMU-Mocap \cite{CMU-Mocap}, AMASS \cite{Mahmood2019AMASSAO} and 3DPW \cite{vonMarcard2018}. The Human3.6M dataset \cite{ionescu2013human3} is a massive motion capture dataset comprising over 3.6 million human poses and corresponding high-resolution images captured using a high-speed motion capture system. The CMU-Mocap dataset \cite{CMU-Mocap} is a widely used dataset for human pose prediction, containing 8 action categories and 600K frames (38 keypoints). AMASS dataset \cite{Mahmood2019AMASSAO} is significantly richer than previous human motion collections, with more than 40 hours of motion data and more than 11k motions. 3DPW dataset \cite{vonMarcard2018} includes 60 sequences, more than 51k frames, and 7 actors, and uses 9-10 IMUs per person to simultaneously track up to 2 subjects. The above datasets are commonly-used benchmarks for single-person motion prediction. To integrate scene context, Cao \etal \cite{caoHMP2020} propose a GTA Indoor Motion dataset (GTA-IM) that emphasizes human-scene interactions in indoor environments. Zheng \etal \cite{Zheng2022GIMOGH} propose a GIMO dataset that delivers high-quality body pose sequences, scene scans, as well as ego-centric views with eye gaze that serves as a surrogate for inferring human intent.

\textbf{Multi-person based datasets.} Apart from the above datasets, several works have evaluated their multi-person motion prediction models using the 3DPW \cite{vonMarcard2018} and MuPoTs-3D \cite{Mehta2017SingleShotM3} datasets. The MuPoTs-3D dataset \cite{Mehta2017SingleShotM3} is initially designed for pose estimation and consists of over 8k frames from 20 real-world scenes featuring 2$\sim$3 subjects and 14 keypoints.  Furthermore, the CMU-Mocap dataset \cite{CMU-Mocap} also includes 2 interactive subjects while the scale of this dataset is relatively small, with only 3k frames available. To advance research in multi-person motion prediction, Guo \etal \cite{guo2022multi} capture poses from two professional dancers and introduced the EXPI dataset, which consists of 30k frames with 60k annotations and 18 keypoints. Despite its usefulness, the EXPI dataset \cite{guo2022multi} has some limitations in its ability to model more complex scenarios, as it is designed to only capture interactions between two subjects. While this restriction may be appropriate for some applications, it may not be sufficient for others that require the modeling interactions of more individuals in a given scene. In this paper, we propose a large-scale Multiple-person Interaction Motion (MI-Motion) dataset that contains 5 different daily activity scenes and collects 167k pose frames with 3$\sim$6 interactive subjects.

\subsection{Algorithms and Benchmarks for 3D Human Motion Prediction}

\textbf{Single-person motion prediction.} Recent years have witnessed a bunch of work for single-person motion prediction \cite{Tang2018LongTermHM, Mao2019LearningTD, HernandezRuiz2018HumanMP, Diller2020ForecastingC3, cao2020long, kundu2019bihmp, Liu2021MotionPU, Katircioglu2021DyadicHM, Wang2021SimpleBF}. Traditional methods often utilize recurrent neural networks (RNNs) \cite{Graves2013GeneratingSW, mikolov2010recurrent} to address this sequence-to-sequence problem but still suffer in error accumulation \cite{martinez2017human, Fragkiadaki2015RecurrentNM}. To solve this issue, feed-forward networks, such as graph convolution networks (GCNs) \cite{Zhang2019GraphCN} and temporal convolution networks (TCNs) \cite{Lea2016TemporalCN}, are utilized to capture spatial and temporal dependencies \cite{Zhong2022SpatioTemporalGG, Dang2021MSRGCNMR, Sofianos2021SpaceTimeSeparableGC, Li2021SkeletonGS}. For instance,  Li \etal \cite{li2020dynamic} designed a novel GCN named DMGNN which contained a dynamic multi-scale graph to represent the human skeleton structure. STS-GCN \cite{Sofianos2021SpaceTimeSeparableGC} is the first space-time-separable GCN, whose space-time graph connectivity is factored into space and time affinity matrices. It bottlenecks the space-time cross-talk while enabling full joint-joint and time-time correlations. Besides, some attention-based methods are naturally introduced to model pose dynamics. In particular, Mao \etal \cite{mao2020history} propose an attention-based network that captures the similarity between the current motion context and historical motion sub-sequences. Aksan \etal \cite{aksan2021spatio} propose a spatial-temporal transformer that learns high-dimensional embeddings for skeletal joints and generates temporally coherent poses via a decoupled temporal and spatial self-attention mechanism. The majority of these single-person based methods in the literature follow consistent training settings, whereby 10 frames are inputted as observed sequence and output 10 frames and 25 frames as short-term prediction ($<400 ms$) and long-term prediction ($400 ms\sim1000 ms $), respectively.

\textbf{Multi-person motion prediction.} The above-mentioned methods model pose dynamics for each individual in isolation, ignoring complex human-human body interaction. To solve this issue, recent approaches are proposed for multi-person pose and trajectory forecasting (\aka multi-person motion prediction). For example, Adeli \etal \cite{adeli2021tripod} introduce to combine scene context and use graph attention networks to model the both human-human and human-object interactions in 2D scenarios. Wang \etal \cite{wang2021multi} first propose 3D multi-person motion prediction task and presented local-global transformers to model individual motion and social interactions, respectively. Guo \etal \cite{guo2022multi} design a two-branch attention network to model two interacted persons via cross-attention mechanism. For these works, skeletal body interaction between individuals is not captured effectively. Therefore, Peng \etal \cite{peng2023trajectory} propose a transformer-based framework called TBIFormer that separates human body parts based on body semantics and effectively learn body parts interactions for intra- and inter-individuals. Vendrow \etal \cite{vendrow2022somoformer} present SoMoFormer that leverages a joint sequence representation for human motion input instead of a time sequence, enabling joint-level attention mechanisms while simultaneously predicting entire future motion sequences in parallel for each joint. Currently, the lack of a unified benchmark for training and evaluation poses a challenge in the multi-person motion prediction, which hinders fair comparisons and limits progress. Hence, we address this gap and establish a comprehensive benchmark for training and evaluation to foster future research.

\vspace{-2mm}
\section{The MI-Motion Dataset}
\vspace{-2mm}
In this section, we introduce the proposed MI-Motion dataset, which emphasizes collecting three or more interactive subjects to provide more complex and diverse interaction scenes. The comparison between our dataset and other available datasets are illustrated in Table~\ref{tab:dataset}.
\subsection{Dataset Overview}
As shown in Figure \ref{fig:teaser}, the MI-Motion dataset consists of 5 daily activity scenes: indoor, park, street, special locations and complex crowd. Specifically, the indoor scene primarily encompasses motions occurring within a limited range of regions, such as sitting, wandering, and indoor work. The park scene involves various activities, including romantic behaviors, playing, dancing, and more. The street scene includes motions like walking, talking, running, and cycling at a fast pace. The special locations scene showcases less common motions, such as extreme sports, combat, and outdoor labor. Finally, the complex crowd scene combines all the aforementioned motion types to create a comprehensive representation. In summary, we produced a total of 217 sequences using over three hundred movements, including 201 three-person sequences and 16 six-person sequences, and we ended up with 167K sequence frames, each including the exact global position of 20 keypoints for each subject.
\subsection{Data Collection, Preprocessing and Synthesis}
During the data collection phase, we initially gathered and purchased a substantial number of interactive action packs from the open-source community and Unreal Engine Marketplace\footnote{https://www.unrealengine.com/marketplace/en-US/store}, respectively. These action packs encompass a wide range of interactive and non-interactive animation sequences, typically stored in the Asset or FBX format. In addition, to collect some extreme and special motions that are not available in open resource, we utilize a marker-based motion capture system\footnote{https://www.rokoko.com/} to record them into BVH files. By utilizing these action packs and captured motion sequences as motion templates, we can further finetune them for refinement, ultimately enabling us to construct a diverse array of interactive scenes during the synthesis phase.

To streamline the synthesis process, we preprocess the motion sequences into a unified format. Initially, we bind all the motion sequences into a standardized skeletal mesh termed SK\_Mannequin within the Unreal Engine 5. Each skeletal body comprises 20 skeleton joints. Subsequently, we develop a Blueprint that precisely captures the position of each skeleton joint and exports the data into the desired file format. More specifically, we utilize the Blueprint to access the skeletal meshes within the current scene. Using a built-in API, we extract the global positions of the designated body joints for each frame and save all of them into JSON files.

Upon completion of the aforementioned preprocessing, we are now poised to customize five activity scenes tailored to our specific requirements. We meticulously select the motion templates and construct motion groups for the scenes of \textit{Park} (3 subjects), \textit{Street} (3 subjects), \textit{Indoor} (3 subjects), \textit{Special Locations} (3 subjects) and \textit{Complex Crowd} (6 subjects), featuring three levels of interaction, as illustrated in Figure \ref{fig:interact_level}-(b). This enables us to simulate diverse possibilities in real-life scenarios. Specifically, for each motion group, we create a Level Sequence\footnote{https://docs.unrealengine.com/5.0/en-US/unreal-engine-sequencer-movie-tool-overview/} to govern the animations. We incorporate the selected motion templates into the animation track, finetuning their initial position and motion trajectory of each skeletal mesh by manipulating the transform curve of each character within the track. Additionally, when necessary, we utilize the animation editor to refine the motion group, ensuring a more natural and realistic interaction effect. Finally, we execute the debugged composite animation and employ the active blueprint to output the global position coordinates of each skeletal mesh's keypoints.

\subsection{Dataset Structure, Storage and Access}
Our full dataset contains over 167k frames and almost multi-person pose sequences. More precisely, for each recorded sequence MI-Motion provides:
\begin{itemize}[
    topsep=0pt,
    leftmargin=*]
 \item Pose sequences at 75 FPS. For training, we downsample them to 25 FPS as previous methods do \cite{guo2022multi, mao2020history}.
  \item Motion data: 3D position of all joints per frame. 20 joints of the all subjects are tracked by Unreal Engine at default. For simplicity of training, we select 18 primary joints (one in the head, one in the neck, four in the spines, and both shoulders, elbows, wrists, hips, knees, heels), as shown in Table~\ref{tab:KeypointDefinition} and Figure~\ref{fig:KeypointDefinition} of Appendix~\ref{app:dataset}. We provide a PYTHON script for the above operations.
\end{itemize}

We extract data from JSON files and utilize PYTHON to store them in a standard geometric Numpy file format (\ie .npy). More details about the dataset are attached in Appendix~\ref{app:dataset}. Besides, we develop a website\footnote{\href{https://mi-motion.github.io/}{\texttt{https://mi-motion.github.io/}}} to host our dataset and provide more details and experiment results.

\section{Benchmark for 3D Multi-Person Motion Prediction}
In this section, we introduce a comprehensive benchmark for 3D Multi-Person Motion Prediction using the MI-Motion dataset. We evaluate and compare three prediction models, including one representative single-person prediction model and two recently-released multi-person prediction models. Our evaluation employs a unified standard to ensure fair comparisons and includes performance analysis for each method. By establishing this benchmark, we aim to facilitate advancements in the field of multi-person motion prediction and foster the development of more effective and accurate prediction models.

\vspace{-2mm}
\paragraph{Problem Definition.} In this context, we introduce the multi-person motion prediction task, where the objective is to evaluate the ability of a model to accurately and smoothly predict poses and trajectories for multiple individuals involved in various interactions. Supposing the observed skeletal poses from person $p$ are $X_{1:T}^p=\{x_1^p,x_2^p,...,$ $x_{T+1}^p\}$ with $T$ frames, where $p = 1,2, ... P$. Our goal is to predict the $N$ frames of future poses and trajectories $X_{T+1:T+N}$ for all the individuals.

\vspace{-2mm}
\paragraph{Baseline Methods.} To aid future research, we benchmark three code-released baseline algorithms from the literature. The one is for classical single-person motion prediction: HRI \cite{mao2020history},  and the two for multi-person settings: MRT \cite{wang2021multi}, TBIFormer \cite{peng2023trajectory}. The above methods are attention or transformer \cite{vaswani2017attention} based methods. In this work, we propose a lightweight and efficient pure convolutional method called Social Temporal GCN (SocialTGCN). SocialTGCN consists of three main components: a Pose Refine Module (PSM), a Social Temporal GCN (SocialTGCN) encoder, and a TCN decoder. Detailed information about all the baselines, as well as ablation results for the proposed SocialTGCN, can be found in Appendix~\ref{app:baseline-models}. 

\subsection{Evaluation for 3D Multi-Person Motion Prediction}
\subsubsection{Evaluation Settings}
Following most single-person based methods \cite{mao2020history, Dang2021MSRGCNMR, Sofianos2021SpaceTimeSeparableGC}, we also evaluate predicted results on short-term ($80 \sim 400ms$) and long-term ($400 \sim 1000ms$). Specifically, we input the model with 25 frames ($1000ms$) and forecast 25 frames of pose. To further evaluate the performance of the model on longer time horizons, we utilize the pretrained model to autoregressively forecast an additional 25 frames, extending the prediction horizon to $2000ms$, which we term as ultra-long-term prediction. 

\subsubsection{Evaluation Metrics} 
We propose two metrics to measures motion and pose error under multi-person setting, Global Joint Position Error (GJPE), Aligned Joint Position Error (AJPE), and one metric for global trajectory movement, termed body Root joint final Displacement Error (RFDE). These three metrics are defined as below.

\vspace{-2mm}
\paragraph{GJPE metric.} Based by commonly used evaluation metric Mean Per Joint Position Error (MPJPE) \cite{ionescu2013human3} for single-person motion prediction, we modify it to measure both the body poses and global trajectories of all the individuals:

\begin{table}[t] 
\begin{center}
  \caption{
  \textbf{Short-term prediction performance on MI-Motion dataset for 5 different scenes.} The error is evaluated by the three metrics in millimeter. * denotes a classical single-person prediction method. Column-wise best result in bold, second best underlined.}
  \label{tab:short_term}
  \resizebox{\linewidth}{!}
  {
  \begin{tabular} {cl aaaa cccc aaaa cccc aaaa}
    \multicolumn{2}{c}{}&\multicolumn{4}{c}{\cellcolor{SceneBlue}\textit{Park}} & \multicolumn{4}{c}{\cellcolor{SceneRed}\textit{Street}} & \multicolumn{4}{c}{\cellcolor{SceneOrange}\textit{Indoor}} & \multicolumn{4}{c}{\cellcolor{ScenePurple}\textit{Special Locations}} & \multicolumn{4}{c}{\cellcolor{SceneGreen}\textit{Complex Crowd}}\\ \toprule 
    \multicolumn{2}{c}{Time (ms)} & 80 & 160 & 320 & 400 & 80 & 160 & 320 & 400  & 80 & 160 & 320 & 400 & 80 & 160 & 320 & 400 & 80 & 160 & 320 & 400 \\  \midrule
    \multirow{4}{*}{\rotatebox{90}{\textbf{GJPE}}} &  HRI* \cite{mao2020history} & \multicolumn{1}{a}{58} & 48 & 75 & 87 & 52 & 44 & 73 & 88 & 48 & \underline{45} & 78 & 91 & 73& 93& 163& \underline{193}& 43& 48& 82& 95 \\ 
     &MRT \cite{wang2021multi} &  \multicolumn{1}{a}{23} & 44 & 76 & 88 & \underline{20} & 39 & 74 & 89 & \underline{25} & 50 & 80 &101  &47 &90 &\underline{159} &\textbf{189} &24 &47 &88 &106\\ 
     &TBIFormer \cite{peng2023trajectory} &  \multicolumn{1}{a}{\underline{21}} &\underline{36} & \underline{64} &\underline{75}  & \underline{20} & \underline{33} & \underline{60} & \underline{74} & \textbf{20} & \textbf{37} & \underline{69} &\underline{84} &\textbf{35} &\textbf{80} &\textbf{158} &\textbf{189} &\textbf{18} &\textbf{32} &\textbf{63} &\textbf{78}\\ 
     &SocialTGCN &  \multicolumn{1}{a}{\textbf{18}} & \textbf{34} & \textbf{60} & \textbf{72}   & \textbf{15} & \textbf{28} & \textbf{54} & \textbf{64}   & \textbf{20} & \textbf{37} & \textbf{67} & \textbf{81}    &\underline{45} &\underline{89} &165 &199   &\underline{20} &\underline{37} &\underline{70} &\underline{85} \\ 
    \midrule
    \multirow{4}{*}{\rotatebox{90}{\textbf{AJPE}}} &  HRI* \cite{mao2020history} & \multicolumn{1}{a}{20} & 35 & 61 & 68 & 19 & 32 & 56 & 66 & 19 & 35 & 68 & 80 & 41 & 77 & 127 & 145 & 19 & 35 & 63 & 72 \\ 
     &MRT \cite{wang2021multi} &  \multicolumn{1}{a}{20} & 37 & 63 & 72 & 15 &28 &52 &62 & 21 & 41 &73 & 84 &41 &77 &124 &140 &16 &31 &57 &67  \\ 
     &TBIFormer \cite{peng2023trajectory} &  \multicolumn{1}{a}{\textbf{10}} & \textbf{24} &\underline{48} &\underline{57} & \textbf{8} & \textbf{19} &\underline{41} & \underline{51} &\textbf{11} & \textbf{26} & \underline{51} &\underline{63}  &\textbf{25} &\textbf{63} &\underline{118} &\underline{136} &\textbf{9} &\textbf{23} &\underline{49} &\underline{60}\\ 
     &SocialTGCN &  \multicolumn{1}{a}{\underline{14}} & \underline{26} & \textbf{45} & \textbf{53}   & \underline{11} & \underline{21} & \textbf{38} & \textbf{46}    & \underline{15} & \underline{27} & \textbf{49} & \textbf{58}   &\underline{36} &\underline{69} &\textbf{117} &\textbf{134}   &\underline{14} &\underline{26} &\textbf{48} &\textbf{56} \\
     \midrule
    \multirow{4}{*}{\rotatebox{90}{\textbf{RFDE}}} &  HRI* \cite{mao2020history} & \multicolumn{1}{a}{56} & \underline{41} & 59 & 71 & 50 & 38 & 63 & 77 & 49 & 47 & \underline{78} & \textbf{89} & 68 & 76 & 137 & 160 & 41 & 41 & 70 & 83 \\ 
     &MRT \cite{wang2021multi} &  \multicolumn{1}{a}{\textbf{17}} &\textbf{32} &57 &68 & \underline{15} & 29 & 52 & 62 & \underline{22} & 44 & 82 &96 & \underline{36} & \textbf{68} & \textbf{122} & \textbf{144}  &20 &40 &78 &96 \\ 
     &TBIFormer \cite{peng2023trajectory} &  \multicolumn{1}{a}{20} & \textbf{32} & \underline{56} & \underline{66} & 18 & \underline{28} &\underline{48} &\underline{58} & \textbf{20} & \underline{39} & 78 & 93 &\textbf{30} &\textbf{68} &\underline{134} &\underline{158} &\textbf{16} &\textbf{27} &\textbf{53} &\textbf{66} \\ 
     &SocialTGCN &  \multicolumn{1}{a}{\underline{18}} & \textbf{32} & \textbf{53} & \textbf{64}    & \textbf{13} & \textbf{23} & \textbf{41} & \textbf{50}    & \textbf{20} & \textbf{36} & \textbf{73} & \underline{90}    &39 & \underline{74} &142 &174   &\underline{18} & \underline{32} &\underline{62} &\underline{77}\\
     \bottomrule
  \end{tabular}
  }
\end{center}
\vspace{-5mm}
\end{table}

\begin{table}[t] 
\begin{center}
  \caption{
  \textbf{Long-term prediction performance on MI-Motion dataset for 5 different scenes.} The error is evaluated by the three metrics in millimeter. * denotes a classical single-person prediction method. Column-wise best result in bold, second best underlined.}
  \label{tab:long_term}
  \resizebox{\linewidth}{!}
  {
  \begin{tabular}{cl aaa ccc aaa ccc aaa}
    \multicolumn{2}{c}{}&\multicolumn{3}{c}{\cellcolor{SceneBlue}\textit{Park}} & \multicolumn{3}{c}{\cellcolor{SceneRed}\textit{Street}} & \multicolumn{3}{c}{\cellcolor{SceneOrange}\textit{Indoor}} & \multicolumn{3}{c}{\cellcolor{ScenePurple}\textit{Special Locations}} & \multicolumn{3}{c}{\cellcolor{SceneGreen}\textit{Complex Crowd}}\\ \toprule 
    \multicolumn{2}{c}{Time (ms)} & 560 & 720 & 1000 & 560 & 720 &1000 & 560 & 720 & 1000 & 560 & 720 &1000 & 560 & 720 & 1000  \\   
    \midrule

    \multirow{4}{*}{\rotatebox{90}{\textbf{GJPE}}} &  HRI* \cite{mao2020history} & 136 & 150 & 214 & 123 & 126 & 181 & 133 & 140 &195 & 241 & \underline{254} & 331 & 130 & 139 & 206 \\ 
     &MRT \cite{wang2021multi} &  {107} & 124 & \underline{149} & 113 &130 &151 & \underline{119} & 132 &\textbf{147} & \textbf{225} &\textbf{250}  & \textbf{289} & 140 & 170 & 220  \\ 
     &TBIFormer \cite{peng2023trajectory} &  \underline{96} & \textbf{114} & \textbf{141} & \underline{96} & \underline{111} & \underline{131} & \textbf{108} & \underline{129} &\underline{154} & \underline{236} &269 & \underline{312} & \textbf{104} & \textbf{125} &\textbf{158}  \\ 
     &SocialTGCN &  \textbf{95} & \underline{116} & 154 & \textbf{81} & \textbf{98} & \textbf{124} & \textbf{108} & \textbf{127} &160 &246 &276 &322 &\underline{113} & \underline{137} & \underline{177}  \\ 
    \midrule
    \multirow{4}{*}{\rotatebox{90}{\textbf{AJPE}}} &  HRI* \cite{mao2020history} & {\underline{71}} & \underline{80} & 95 & \underline{64} & \underline{72} & \textbf{73} & 85 & 96 &109 & \textbf{149} & \underline{160} & 173 & \underline{72} & \underline{83} & \textbf{94}  \\ 
     &MRT \cite{wang2021multi} & 85 & 92 & 103 & 78 &84 &84 & 97 & 102 &\underline{105} & \underline{154} & \textbf{159} & \underline{171} &82 & 94 & 110   \\ 
     &TBIFormer \cite{peng2023trajectory} & \underline{71} & 81 & \underline{94} & 66 & 74 &\underline{74} & \underline{81} &\underline{95} &\underline{105} & \underline{154} &\textbf{159} &\textbf{169} & 76 &87 &101  \\ 
     &SocialTGCN &  \textbf{66} & \textbf{76} & \textbf{93} & \textbf{58} & \textbf{66} &76 & \textbf{73} & \textbf{83} &\textbf{97} & \underline{154} & 161 & 174 & \textbf{70} & \textbf{81} & \underline{97} \\
     \midrule
    \multirow{4}{*}{\rotatebox{90}{\textbf{RFDE}}} &  HRI* \cite{mao2020history} & {134} & 148 & 213 & 110 & 115 & 170 & 152 & 155 & 221 & 214 & \underline{235} & 313 & 128 & 138 & 211   \\ 
     &MRT \cite{wang2021multi} & \underline{88} &\underline{108} &\underline{142} & 80 &98 &127 &\underline{121} & \underline{143} &\textbf{169} &\textbf{181} & \textbf{215} &\textbf{262} & 131 & 162 &216   \\ 
     &TBIFormer \cite{peng2023trajectory} &  {\textbf{86}} &\textbf{106} &\textbf{136} & \underline{75} & \underline{91} &\underline{117} & \textbf{119} & \underline{143} &\underline{174} & \underline{208} & 251 & \underline{294} & \textbf{90} &\textbf{111} & \textbf{149}  \\ 
     &SocialTGCN & \underline{88} & 112 & 154 & \textbf{62} & \textbf{79} & \textbf{108} & \textbf{119} & \textbf{138} &180 & 226 & 265 & 321 & \underline{106} & \underline{131} & \underline{173} \\
     \bottomrule 
  \end{tabular}
  }
\end{center}
\vspace{-2mm}
\end{table}

\begin{equation}\label{eq1}
\small{
{\rm GJPE}(X,\hat{X})=\frac{1}{P\times J}\sum_{i=1}^{P}\sum_{j=1}^{J}{||X_j^i - \hat{X}_j^i||^2},
}
\end{equation}
where $P$ and $J$ are the numbers of people and body joint. $X_j^i$ and $\hat{X}_j^i$ are the estimated and ground-truth positions of the joint $j$ for person $i$.

\paragraph{AJPE metric.} We remove global movement and use Aligned mean per Joint Position Error (AJPE) to measure pure pose position error:
\vspace{-2mm}
\begin{equation}\label{eq2}
\small{
{\rm AJPE}(X,\hat{X})=\frac{1}{P\times J}\sum_{i=1}^{P}\sum_{j=1}^{J}{||(X_j^i -X_{r}^i) - (\hat{X}_j^i-\hat{X}_r^i)||^2},
}
\end{equation}
\vspace{-3mm}
where $X_{r}^i$ and $\hat{X}_{r}^i$ are the estimated and ground-truth body root positions for person $i$.

\paragraph{RFDE metric.} We introduce Root joint Final Displacement Error (RFDE) that adopts the body root position to evaluat the global movement for all the individuals using a typical trajectory prediction metric: Final Displacement Error. The formula is described as follows:
\vspace{-2mm}
\begin{equation}\label{eq3}
\small{
{\rm RFDE}(X_{r},\hat{X}_{r}) = \frac{1}{P}\sum_{i=1}^{P}{||X_{r,N}^i - \hat{X}_{r,N}^i||^2},
}
\end{equation}

\vspace{-3mm}
where $X_{r,N}^i$ and $\hat{X}_{r,N}^i$ are the estimated and ground-truth root position of final pose at $N$-th predicted timestamp for person $i$.

\begin{figure}[t]
  \centering
  \includegraphics[width=1.0\linewidth]{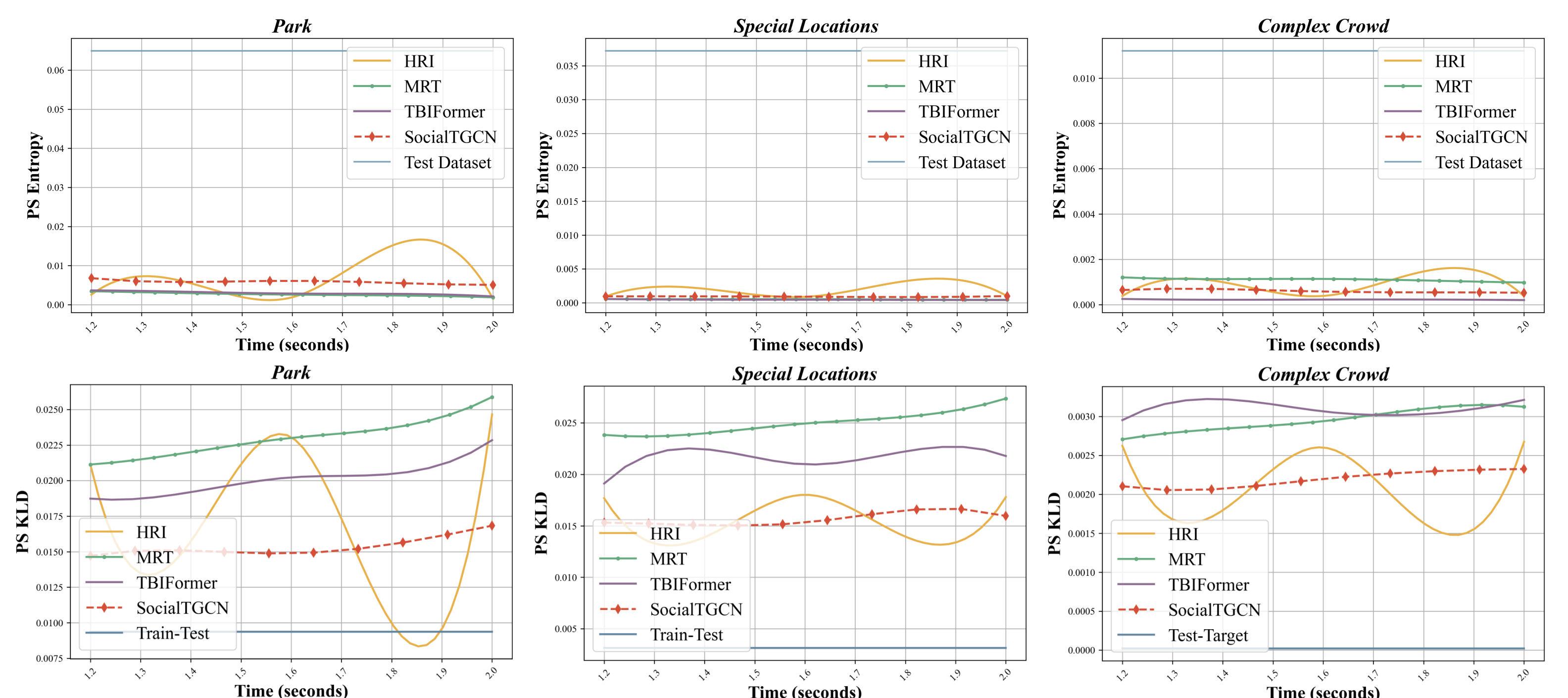}
  \caption{
  \textbf{Power Spectrum (PS) metrics.} (Top) PS Entropy: A higher value indicates better performance. Our model (shown in red, dashed line) demonstrates higher entropy, indicating its resilience to the static pose problem during longer predictions. (Bottom) PS KLD: A lower value indicates better performance. Our model consistently maintains a prediction distribution that closely aligns with the data distribution, as evidenced by the symmetric KLD.}
  \label{fig:ps_kld}
  \vspace{-5mm}
\end{figure}

\paragraph{Power Spectrum (PS) metrics.} In multi-person motion prediction, the generation of global motion for each individual is essential. While the baselines ensure local consistency in many cases, there may be instances where the global positions deviate from the ground truth due to natural variations \cite{aksan2021spatio}. To assess the quality of predictions in multi-person motion for longer horizon, we adopt distribution-based metrics in the power spectrum (PS) domain, proposed as an alternative to direct ground-truth comparisons \cite{Hernandez_Gall_Moreno_2020}. Specifically, we introduce two metrics: (i) PS KLD, quantifying the dissimilarity between prediction and test distributions using the KL divergence, and (ii) PS Entropy, capturing the diversity of predictions in the power spectrum. The latter metric evaluates the degree to which predictions deviate from a static pose, with lower entropy indicating a more limited range of predictions. However, PS Entropy should be interpreted alongside PS KLD, as it can be influenced by random predictions \cite{Hernandez_Gall_Moreno_2020}. Only predictions resembling the real data samples can achieve a lower score in PS KLD, ensuring a meaningful evaluation. The detailed definitions of the metrics can be found in Appendix~\ref{app:evaluation-metrics}.

\vspace{-2mm}
\section{Experiments with MI-Motion}
\label{sec:experiments}
\vspace{-2mm}
In this section, we study the performance of the existing multi-person motion prediction models in the three evaluation settings.

\vspace{-2mm}
\paragraph{Implementation details.} In each experiment, we use 80\% data from the scenes (\textit{Park, Street, Indoor} and \textit{Special Locations}) for training and the remaining 20\% for testing. Besides, the data from \textit{Complex Crowd} are all for testing. We sample 75 frames from each motion sequence at a frame rate of 25 frames per second (FPS). Out of these frames, 25 frames are utilized as input, while the remaining 50 frames serve as ground truth for prediction evaluation. More training and implementation details are summarized in Appendix~\ref{app:implementation}.

\begin{wrapfigure}{t}{0.45\linewidth}
\makeatletter\def\@captype{table}\makeatother
\vspace{-0.2cm}
\centering
\caption{\textbf{Comparative results of model parameters and FLOPs.}}
\label{tab:model_capacity}
  \resizebox{0.9\linewidth}{!}
  {
    \begin{tabular}{lcc}
    \toprule
    Method  & Params (M) & FLOPs (G)  \\
    \midrule
    HRI \cite{mao2020history}            & \textbf{2.83} &2.829 \\
    MRT \cite{wang2021multi}             & 7.28         & \underline{0.382}  \\
    TBIFormer \cite{peng2023trajectory}  & 5.60         & 0.597 \\
    SocialTGCN                           & \underline{3.31}          & \textbf{0.320}\\ \bottomrule
    \end{tabular}
    }
\vspace{-0.4cm}
\end{wrapfigure}

\vspace{-2mm}
\paragraph{Quantitative analysis.} Table~\ref{tab:short_term} and Table~\ref{tab:long_term} report the results of GJPE, AJPE and RFDE on the 5 different scenes for short-term and long-term prediction, respectively. HRI \cite{mao2020history}, being a single-person based method focusing on local pose dynamics, exhibits subpar performance in both GJPE and RFDE metrics. This can be attributed to the absence of a suitable loss constraint tailored for its learning objectives on global motion dynamics. MRT \cite{wang2021multi} and TBIFormer \cite{peng2023trajectory} are transformer-based methods designed for multi-person scenarios, where TBIFormer \cite{peng2023trajectory} performs better than MRT \cite{wang2021multi} due to its body interaction modeling and spatial encoding strategy. Furthermore, our proposed baseline, SocialTGCN, also achieves competitive results on both short-term and long-term horizons, while maintaining a lightweight and low complexity, as illustrated in Table~\ref{tab:model_capacity}.

For ultra-long-term prediction, we report the Power Spectrum (PS) results in Figure~\ref{fig:ps_kld}. Our proposed SocialTGCN shows better performance in PS Entropy and PS KLD. Notably, HRI \cite{mao2020history} often produces unsmooth motions characterized by jittering and instant movement. Cause that these phenomena is not easy to show in the static plots, we have included GIFs on the dataset website. We believe that the presence of these generated unnatural movements may result in substantial fluctuations in the power spectrum (PS) plots. As a result, the PS results of the HRI method may not be suitable for reference, despite exhibiting better values during certain periods in the plots.

\begin{figure}[t]
  \centering
  \includegraphics[width=1.0\linewidth]{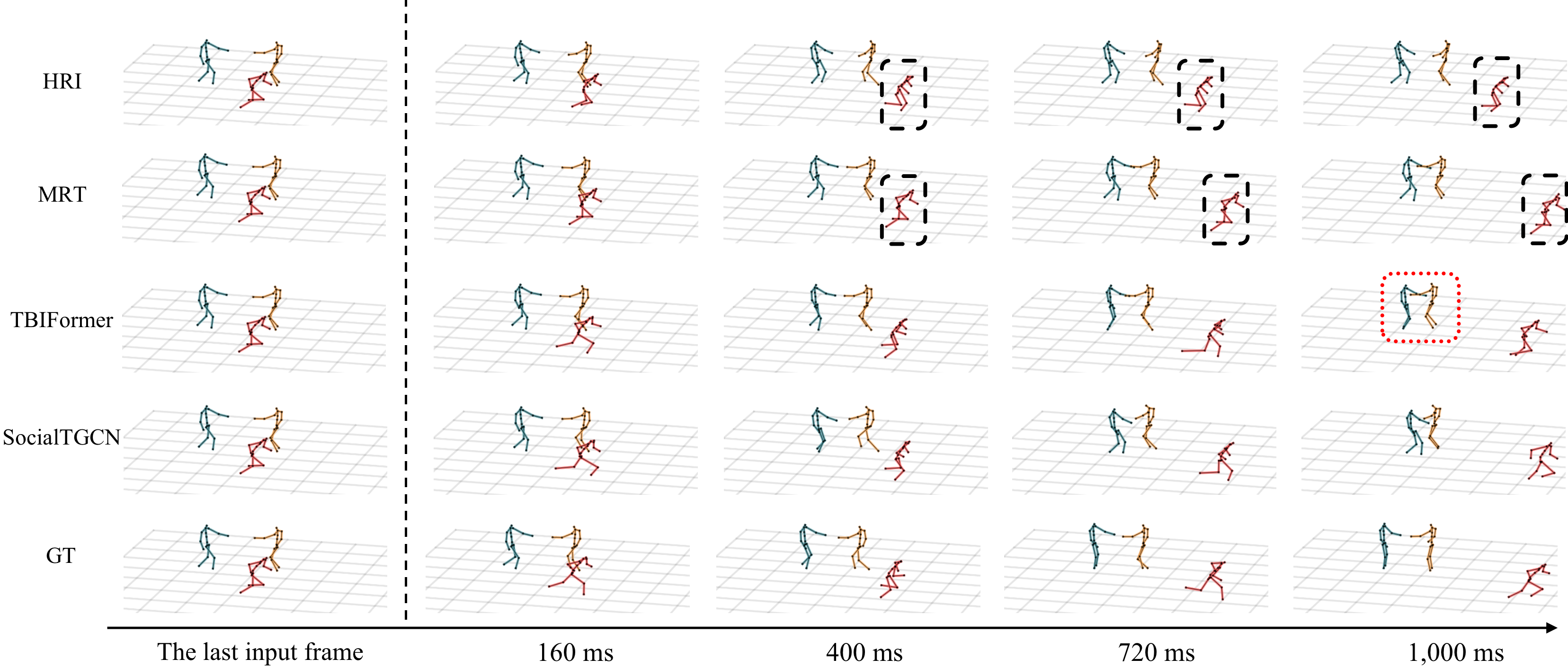}
  \caption{
  \textbf{Qualitative visualization on MI-Motion dataset.} Comparison with the baselines and the ground truth on a sample of the \textit{Park} scene. The left one column are inputs, and the right four columns are predictions.
  }
  \label{fig:qualitative_park}
\end{figure}

\paragraph{Qualitative analysis.}
 Figure~\ref{fig:qualitative_park} shows some visualization results in the \textit{Park} scene of the ground truth and all the baselines for both short-term and long-term prediction. HRI \cite{mao2020history} often produces instant displacement or jittering motion, which may not clear in the static plots. MRT \cite{wang2021multi} demonstrate a tendency to converge towards a static pose in the long-term predictions, as indicated by the black dashed line. In contrast, the TBIFormer and SocialTGCN methods generate more dynamic and plausible motion in practice, exhibiting results that are much closer to the ground truth compared to other methods. Additional visualization results can be found in Appendix~\ref{app:qualitative}. For ultra-long-term prediction, we provide GIFs on the dataset website to facilitate better visualization and comparison. 

\section{Conclusion}
\label{sec:conclusion}
\vspace{-4mm}
In this paper, we present the MI-Motion dataset, a dedicated resource tailored to the multi-person motion prediction task. MI-Motion encompasses interactions among 3 to 6 subjects in diverse daily activity scenes, capturing varying levels of interaction complexity. Additionally, we present SocialTGCN, a novel and efficient convolution-based baseline method that serves as a catalyst for further advancements in this domain. For fair comparison, we benchmark three state-of-the-art and code-released methods alongside SocialTGCN on the MI-Motion dataset, evaluating the performance across three distinct prediction settings. We hope the obtained comprehensive results will provide valuable insights for research and innovation in the realm of multi-person motion prediction.
\vspace{-2mm}
\paragraph{Limitations and future work.}
The scale of the MI-Motion dataset is relatively limited compared to available motion datasets focused on single-person motion. This limitation arises from the inherent difficulty in capturing a large number of samples with multiple individuals engaged in interactions.

In future work, our objectives include designing and constructing a multi-view motion capture system to expand the MI-Motion dataset, encompassing a larger number of subjects with a wider range of interaction types. Additionally, we recognize the increasing importance of scene context in understanding human motion and interaction. To address this, we plan to provide 3D scene information, such as point clouds, to facilitate the study of multi-person motion prediction with multimodal information. These advancements will contribute to a more comprehensive understanding of human motion and enable further progress in the field.

\clearpage
{
\small
\bibliographystyle{plain}
\bibliography{neurips_data_2023}
}

\clearpage
\section*{Checklist}
\begin{enumerate}

\item For all authors...
\begin{enumerate}
  \item Do the main claims made in the abstract and introduction accurately reflect the paper's contributions and scope?
    \answerYes{}
  \item Did you describe the limitations of your work?
    \answerYes{See Section~\ref{sec:conclusion}.}
  \item Did you discuss any potential negative societal impacts of your work?
    \answerYes{See Appendix~\ref{app:broader-impact}.}
  \item Have you read the ethics review guidelines and ensured that your paper conforms to them?
    \answerYes{}
\end{enumerate}

\item If you are including theoretical results...
\begin{enumerate}
  \item Did you state the full set of assumptions of all theoretical results?
    \answerNA{}
	\item Did you include complete proofs of all theoretical results?
    \answerNA{}
\end{enumerate}

\item If you ran experiments (e.g. for benchmarks)...
\begin{enumerate}
  \item Did you include the code, data, and instructions needed to reproduce the main experimental results (either in the supplemental material or as a URL)?
    \answerYes{\href{https://mi-motion.github.io/}{\texttt{https://mi-motion.github.io/}} contains instructions to download and process the data. We also provide the \href{https://github.com/xiaogangpeng/SocialTGCN}{\texttt{SocialTGCN}} repository on GitHub, which contains instructions and codes to reproduce the benchmark results.}
  \item Did you specify all the training details (e.g., data splits, hyperparameters, how they were chosen)?
    \answerYes{See Section~\ref{sec:experiments} and Appendix~\ref{app:implementation}.}
	\item Did you report error bars (e.g., with respect to the random seed after running experiments multiple times)?
    \answerNo{We do not report the error bars. To ensure the robustness and reliability of our experimental results, we performed multiple runs of the experiments using different random seeds. Specifically, we conducted three runs for each experiment and then averaged the results. We observed that the results across different runs were similar, indicating consistency and a small standard deviation.}
	\item Did you include the total amount of compute and the type of resources used (e.g., type of GPUs, internal cluster, or cloud provider)?
    \answerYes{See Appendix~\ref{app:implementation}.}
\end{enumerate}

\item If you are using existing assets (e.g., code, data, models) or curating/releasing new assets...
\begin{enumerate}
  \item If your work uses existing assets, did you cite the creators?
    \answerYes{Appendix~\ref{app:baseline-models} for the baseline methods we used.}
  \item Did you mention the license of the assets?
    \answerYes{See Section~\ref{app:license}.}
  \item Did you include any new assets either in the supplemental material or as a URL?
    \answerNo{We did not include new assets either in the supplemental material or as a URL.}
  \item Did you discuss whether and how consent was obtained from people whose data you're using/curating?
    \answerNA{}
  \item Did you discuss whether the data you are using/curating contains personally identifiable information or offensive content?
    \answerYes{See Section~\ref{app:broader-impact}.}
\end{enumerate}

\item If you used crowdsourcing or conducted research with human subjects...
\begin{enumerate}
  \item Did you include the full text of instructions given to participants and screenshots, if applicable?
    \answerNA{}
  \item Did you describe any potential participant risks, with links to Institutional Review Board (IRB) approvals, if applicable?
    \answerNA{}
  \item Did you include the estimated hourly wage paid to participants and the total amount spent on participant compensation?
    \answerNA{}
\end{enumerate}

\end{enumerate}

\clearpage

\appendix

\section*{Appendix Overview}

We provide additional details and results to complement the main paper.
Specifically, this document includes the following materials:
\begin{itemize}
  \item Broader impact (Appendix~\ref{app:broader-impact}).
  \item License information (Appendix~\ref{app:license}).
  \item Additional dataset details (Appendix~\ref{app:dataset}).
  \item Details of evaluation metrics (Appendix~\ref{app:evaluation-metrics}).
  \item Details of baseline methods (Appendix~\ref{app:baseline-models}).
  \item Additional training and implementation details (Appendix~\ref{app:implementation}).
  \item Additional quantitative results (Appendix~\ref{app:quantitative}).
  \item Additional qualitative results (Appendix~\ref{app:qualitative}).
  \item Dataset accessibility and long-term preservation plan (Appendix~\ref{app:dataset-accessibility}).
  \item Author statement of responsibility (Appendix~\ref{app:author-statement}).
\end{itemize}

\section{Broader Impact} \label{app:broader-impact}
Human motion prediction is an important task to understand human behavior in the real world. We believe MI-Motion benchmark is an important step towards understanding complex human interactions and predict plausible motions of multiple individuals. Our dataset will facilitate future study in this field, and eventually enable robots and machines to understand human intent and movement. \paragraph{Potential negative societal impacts.} All the motion data no matter from Unreal Engine Marketplace or ourselves are captured from real persons via motion capture system. We acknowledge that identifying a person solely based on their poses and movements remains uncertain. However, in this work, we modify the motion templates in the game engine, which means the original poses and trajectory may change, making it more difficult to identify personal information. Hereby, we reckon that there are no significant risks of human rights violations or security threats associated with it.

\section{License Information}
\label{app:license}
We grant our dataset a non-exclusive, royalty-free right to install the data on computers owned or controlled by you and/or your organization for the purpose of conducting non-commercial scientific research. The use of the data for any other purpose, including commercial use, is strictly prohibited. This includes, but is not limited to, incorporation into commercial products, use in commercial services, or production of commercial artifacts such as 3D models, movies, or video games. We provide the base model code of each baseline method, following the MIT licenses specified in HRI \cite{mao2020history}, MRT \cite{wang2021multi} and TBIFormer \cite{peng2023trajectory}, and released the proposed SocialTGCN under MIT license.

\section{Additional Dataset Details}
\label{app:dataset}
\begin{figure}[h]
  \begin{minipage}[b]{0.4\linewidth} 
    \centering 
    \scriptsize
        \tabcaption{The definition of human body keypoints.} 
        \begin{tabular}{cccc}
            \toprule
            Keypoint     & Definition     & Keypoint  & Definition \\
            \midrule
            1    & Head                   & 10   & Spine 2  \\
            2     & Neck                  & 11  & Spine 3   \\
            3     & Right shoulder        & 12  &Spine 4\\
            4    & Right elbow            & 13   & Right hip  \\
            5     & Right wrist           & 14  & Right kneel   \\
            6     & Left shoulder         & 15  & Right heel\\
            7    & Left elbow             & 16   & Left hip  \\
            8     & Left wrist            & 17  & Left kneel   \\
            9     & Spine 1                & 18  & Left kneel\\
            \bottomrule
        \end{tabular}
    \label{tab:KeypointDefinition} 
  \end{minipage}
  \begin{minipage}[b]{0.55\linewidth} 
    \centering
    \includegraphics[width=0.4\linewidth]{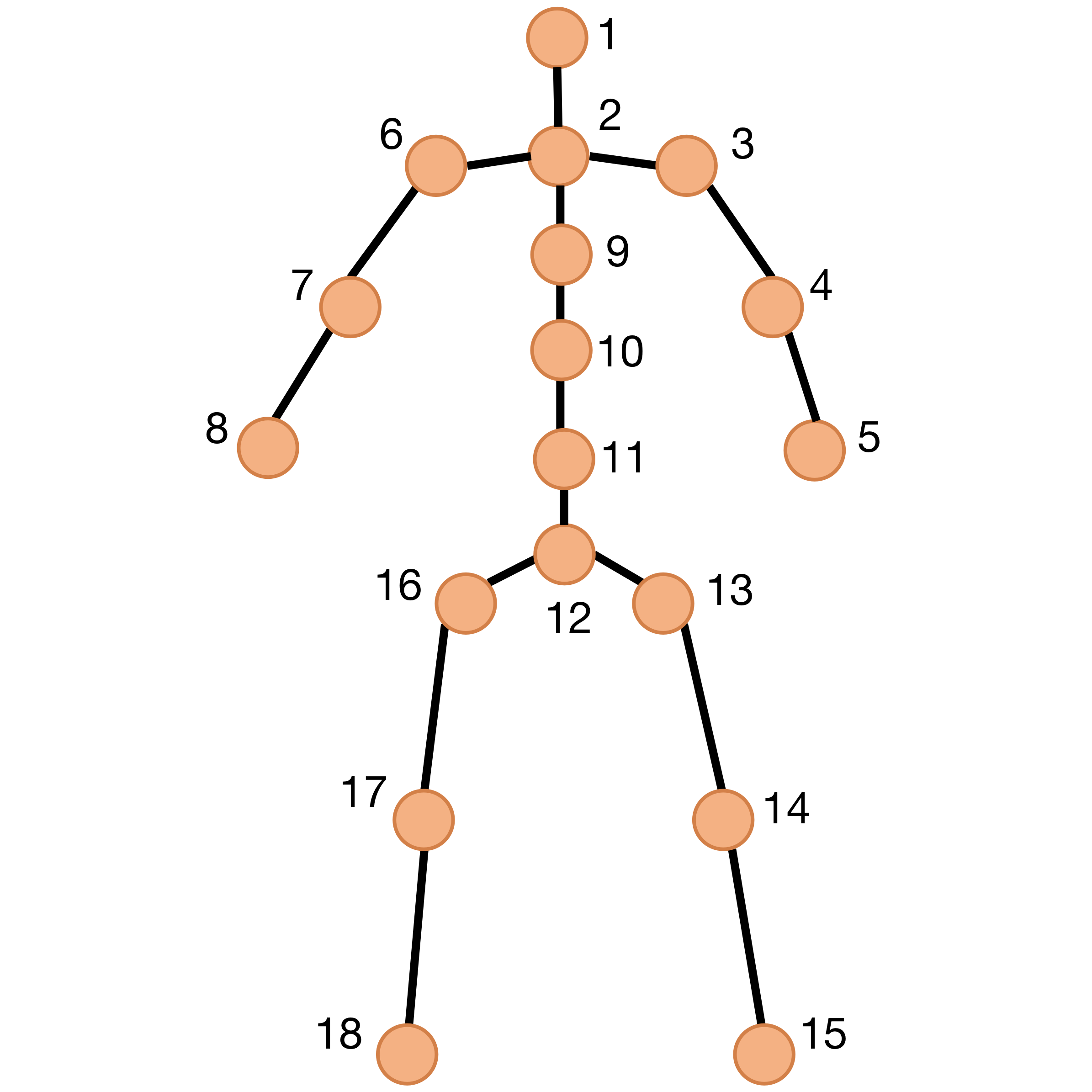}
    \caption{The illustration of human body structure.}
    \label{fig:KeypointDefinition} 
  \end{minipage} 
\end{figure}

\paragraph{Motion data.}
As depicted in Table~\ref{tab:KeypointDefinition} and Figure~\ref{fig:KeypointDefinition}, we provide a definition and illustration of the human body that we construct. Furthermore, in Table~\ref{tab:app-action}, we provide a listing of the representative action content of the motion templates selected for each activity scene.

\newcolumntype{C}[1]{>{\centering\arraybackslash}m{#1}}
\begin{table}[t]
\begin{center}
\caption{The action content of the motion templates synthesized for different scenes encompasses a variety of activities and movements that are representative of the respective scene.}
\label{tab:app-action}
\vspace{2mm}
\begin{tabular}{C{2cm} C{10cm}}
\hline
\textbf{Scene} &  \textbf{Action Content} \\
\hline
\textit{Park} & \small{walking side by side, sitting and talking on the phone, standing and chatting, walking with a cup of coffee, sitting down, two people hugging, proposing marriage, sitting and embracing, kneeling down for a hug, hugging from behind, tiptoeing for a kiss, pointing at the sky while chatting, having a heated argument, etc.}\\
\hline
\textit{Street} & \small{feeling embarrassed after farting, mocking, crossing arms, saluting, standing at attention, promoting a product, browsing items, greeting, repairing things, waving from a distance and running over to meet, mopping the floor, arguing about which direction to go, riding a bicycle (slowly), riding a bicycle (fast), picking up trash, etc.}\\
\hline
\textit{Indoor} & \small{swinging body while drunk, clinking glasses and drinking, shouting to promote, selecting items, mopping the floor, sitting down to order food, standing and serving customers, chatting, eating a hamburger, celebrating at the bar counter, washing dishes, chopping vegetables, using a frying pan, opening the refrigerator, doing yoga, squatting, exercising abdominal muscles, etc.} \\
\hline
\textit{Special Locations} & \small{breaking into through a window, climbing over a railing, somersaulting over a railing sideways, shooting a basketball, playing baseball, doing pull-ups, performing a side flip, boxing, climbing a pole, doing a backflip, doing a forward roll, kicking a door, high-kicking, practicing Chinese martial arts, rowing a boat, practicing taekwondo, riding a motorcycle, pushing a box, walking forward while carrying a box, practicing Jeet Kune Dom, etc.}\\
\hline
\textit{Complex Crowd} & \small{playing the guitar, applauding, swimming freestyle, doing the breaststroke, rowing a boat, engaging in combat, coughing, dodging, retrieving things from the trunk, waving across the street, opening a cabinet to take something, opening a box, checking an oven, flicking a cigarette butt, knocking on a table, contemplating, etc. }\\
\hline
\end{tabular}
\end{center}
\end{table}

\section{Additional Details of Evaluation Metrics}
\label{app:evaluation-metrics}
Using pairwise metrics such as the MPJPE is no longer suitable for long prediction because the ground-truth targets are often much shorter than the prediction horizon and the body global positions mostly deviate from the ground truth due to natural variation. Hence, we use Power Spectrum (PS) metrics proposed by Hernandez \etal \cite{Hernandez_Gall_Moreno_2020} and modified by Aksan \etal \cite{aksan2021spatio} for 3D joint positions, which enables straightforward conversion of any angle-based representation into positions through forward kinematics \cite{aksan2021spatio}. Given a sequence $\vec{X}$, we treat every coordinate of every joint over time as a feature sequence $\vec{x}_f$ following \cite{Hernandez_Gall_Moreno_2020}. The power spectrum PS is then equal to $PS(\vec{x}_f) = ||FFT(\vec{x}_f)||^2$ where $FFT$ denotes the Fast Fourier Transform.

\paragraph{PS Entropy.} The formula is defined as
\begin{equation}
PS~Entropy(\mathcal{X}) = \frac{1}{|\mathcal{X}|}\sum_{\vec{X} \in \mathcal{X}} \frac{1}{F}\sum_{f = 1}^F\sum_{e=1}^E -||PS(\vec{x}_f)|| * \log (||PS(\vec{x}_f)||),
\end{equation}

where $\mathcal{X}$ is either the ground-truth test or training dataset, or the predictions made of a respective model on the corresponding test dataset. $f$ and $e$ correspond to a feature and frequency, respectively.

\paragraph{PS KLD} To compute the PS KLD metric, we sample all the sequences of length $1$ sec (i.e., $25$ frames) from the test dataset and calculate the power spectrum distribution $G$. Then, we get a window of $1$ sec from the ultra-long-term predictions (1.0s $\sim$ 2.0s) to get $P_{t}$ where $t$ stands for the corresponding prediction window of length 1 second. For example, $P_{1}$ is the power spectrum distribution for the predictions between $1$ and $2$ seconds. This approach allows us to assess the accuracy of longer predictions by comparing each individual second of the predicted data with the corresponding real reference data. The symmetric PS KLD metric is then defined as
\begin{equation}
PS~KLD(G, \mathcal{X}, t) = \frac{1}{2|\mathcal{X}|} \sum_{P_t \in \mathcal{X}} KLD(G \mid\mid P_t) + KLD(P_t \mid\mid G).
\label{eqn:sym_ps_kld}
\end{equation}

We employ available implementations of the metrics (\cite{aksan2021spatio}, \href{https://github.com/eth-ait/motion-transformer/blob/master/spl/evaluation.py#L345}{Github link}). It is worth to mention that the PS KLD metric measures the discrepancy between the real and predicted data distributions rather than pairwise comparisons between ground-truth and predictions.

\vspace{-3mm}
\section{Details of Baseline Methods}
\label{app:baseline-models}
\vspace{-2mm}
\subsection{HRI Method}
HRI \cite{mao2020history} is a classical single-person based method that employs an attention-based feed-forward network to explicitly leverage this observation. In particular, instead of modeling frame-wise attention via pose similarity, HRI \cite{mao2020history} propose to extract motion attention to capture the similarity between the current motion context and the historical motion sub-sequences. Then, it aggregates the relevant past motions and processes the result with a graph convolutional network to effectively exploit motion patterns from the long-term history to predict the future poses.
\vspace{-2mm}
\subsection{MRT Method}
MRT \cite{wang2021multi} is the first work for multi-person motion prediction proposed by Wang \etal  They introduce a Multi-Range Transformers (MRT) that utilizes local-range and global-range encoders to model individual motion and social interactions, respectively. Then a transformer decoder is performed for prediction of each person by taking a corresponding pose as a query which attends to both local and global-range encoder features.
\vspace{-2mm}
\subsection{TBIFormer Method}
TBIFormer \cite{peng2023trajectory} is also a transformer based method that learns body parts dynamics for intra- and inter-individuals simultaneously. To provide more accurate spatial information, TBIFormer \cite{peng2023trajectory} introduce a trajectory-aware relative position encoding via measuring movement similarity between multiple individuals.
\vspace{-2mm}
\subsection{Our Proposed SocialTGCN Method}
Following \cite{wang2021multi, peng2023trajectory}, to provide more valuable dynamics information,  we transform observed original pose sequence $X_{1:T}^p=\{x_1^p,x_2^p,...,$ $x_{T}^p\}$ with $T$ frames from person $p$ to displacement sequence $Y_{1:T-1}^p= \{{y}_i\}_{i=1,2,..,T-1}^{p} = \{x_{i+1} - x_{i} \}_{i=1,2,..,T-1}^{p}$, where $p = 1,2, ... P$. It will contain more valuable dynamics information. Given the displacement sequence $Y_{1:T-1}$ of each person, our goal is to predict the $N$ frames of future displacement trajectory $Y_{T:T+N-1}$ and transform it back to the pose space $X_{T+1:T+N}$.

\paragraph{Discrete cosine transformation (DCT).}
Recent works \cite{Zhou_2011, Mao2019LearningTD, mao2020history, wang2021multi, peng2023trajectory} have departed from directly predicting Cartesian coordinates for human motion and instead utilize a Discrete Cosine Transform (DCT) to encode motion into the frequency domain as a set of coefficients. This approach takes advantage of the continuous, smooth, and periodic nature of human motion. By employing DCTs, it becomes possible to predict the complete trajectory for all future frames simultaneously while achieving a more effective representation of human motion.

For a pose diplacement sequence of a single coordinate $(y_1, \ldots , y_{T-1})$, the $l$-th DCT coefficient is computed by
\begin{align}
    C_l = \sqrt{\frac{2}{T-1}}\sum_{t=1}^T \frac{y_t}{\sqrt{1 + \delta_{l1}}}cos\frac{\pi}{2(T-1)}(2t-1)(l-1)
\end{align}
for
\begin{align}
    \delta_{ij} = \begin{cases}
                                   1 & \text{if $i=j$}, \\
                                   0 & \text{if $i\neq j$},
  \end{cases}
\end{align}
where $l = 1, 2, \ldots, T-1$, yielding the same number of coefficients as the sequence length. This process gives the coefficient sequence $(C_1, \ldots , C_{T-1})$ which is used as an input token to the model for the corresponding joints coordinate to predict the final, completed trajectory $(\tilde{C}_1, \ldots , \tilde{C}_{T-1})$. Given this coefficient sequence, we can recover the coordinates using the inverse DCT, where we obtain a specific coordinate by
\begin{align}
    y_t = \sqrt{\frac{2}{T-1}}\sum_{l=1}^{T-1} \frac{\tilde{C}_l}{\sqrt{1 + \delta_{l1}}}cos\frac{\pi}{2(T-1)}(2t-1)(l-1).
\end{align}

\begin{figure}[t]
    \centering
    \includegraphics[width=1.0\linewidth]{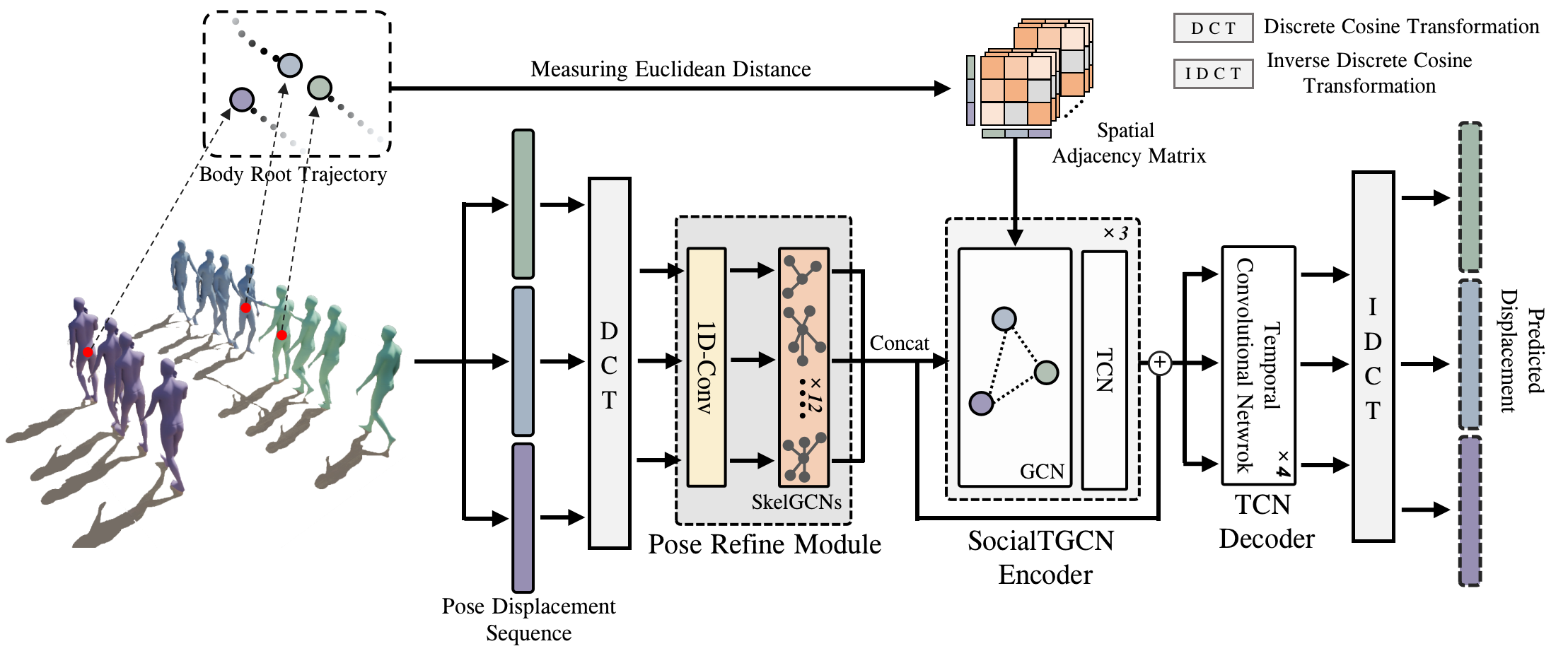}
    \label{fig:pipeline}
    \caption{\textbf{The pipeline of the proposed SocialTGCN.} It mainly consists of three key components: a pose refine module, a socialTGCN encoder, and a TCN decoder.}
\end{figure}

\paragraph{Pose refine module (PSM).}
To capture rich dynamics information, we start by downsampling the displacement sequence of each person using 1D convolutional operators $f$ with a kernel size of $M$ and a stride of $1$, denoted as $\tilde{Y}_{1:T-M} = f(Y)$. This downsampling step helps to retain important dynamics while reducing the sequence length. To refine the pose features of each person and ensure natural and non-distorted predictions, we employ stacked graph convolutional networks (GCNs) with learnable adjacency matrix $\textbf{A}_{skel} \in \mathbb{R}^{3J \times 3J}$. Similar to the one adopted in \cite{mao2020history}, it consists of an initial GCN, an end GCN, and multiple residual GCN blocks, which collectively contribute to improving the quality of the pose features. For person $p$, the output of each GCN block is computed as 
\begin{equation}\label{eq_pose}
\textbf{Z}^{p} = \sigma(\textbf{A}_{skel}^{1}(\sigma(\textbf{A}_{skel}^{0}\tilde{Y}\textbf{W}^{0}))\textbf{W}^{1}),
\end{equation}
where $\sigma(\cdot)$ is tanh activation function, $\textbf{W}$ is weight matrix, and $3J$ denotes 3D coordinates of $J$ body joints.

\paragraph{SocialTGCN encoder.}
We first concatenate (concat) the output $\textbf{Z}^{p} \in \mathbb{R}^{(T-M) \times 3J}$ from the PSM for all the person and have $\textbf{Z} \in \mathbb{R}^{(T-M) \times P \times 3J}$. Then we input it in SocialTGCN encoder, which consists of multiple layers of stacked GCN and TCN. 

Specifically, a GCN composes of a set of graph convolutional layers that are sequentially stacked together and is for modeling social interaction among different individual based on their spatial information. We attach a value $a^{ij}_{spa, t}$ to represent spatial relations between person $i$ and $j$ at time $t$, and we organize $a^{ij}_{spa, t}$s into the weighted adjacency matrix $\textbf{A}_{spa} \in \mathbb{R}^{P \times P}$.  Formally, let $\textbf{Z}^{l} \in \mathbb{R}^{(T-M) \times P \times F^{l}}$ be the input to a graph convolutional layer, $\textbf{A}_{spa} \in \mathbb{R}^{P \times P}$ the spatial adjacency matrix, and $\textbf{W}^{l} \in \mathbb{R}^{F^{l} \times F^{l+1}}$ the trainable parameters, the output of the graph convolutional layer is
\begin{equation}\label{eq_social}
\textbf{Z}^{l+1} = \gamma(\textbf{A}_{spa}^{l}\textbf{Z}^{l}\textbf{W}^{l}),
\end{equation}
where $\textbf{Z}^{l+1} \in \mathbb{R}^{(T-M) \times P \times F^{l+1}}$ and $\gamma(\cdot)$ is leaky relu activation function. The $a^{ij}_{spa, t}$ is computed as 
\begin{equation}
\label{eq_social}
a^{ij}_{spa, t}= \frac{exp(-||x^i_{root, t} - x^j_{root, t}||_2)}{\theta},
\end{equation}
where $x^i_{root, t}$ is body root joint's position of person $i$ at time $t$, and $\theta$ is a scaling factor. For temporal modeling, we adopt a general Temporal Convolutional Networks (TCN) \cite{Gehring_Auli_Grangier_Yarats_Dauphin_2017, bai2018empirical, Luo_Yang_Urtasun_2018} to learn dynamics only on the time dimension.
\paragraph{TCN decoder.} 
Given the encoded observed body dynamics of multiple individuals, the prediction of the 3D coordinates of the body joints in the future is performed by applying convolutional layers to the temporal dimension. These map the observed frames into the future horizon and refine the prediction via a multilayered architecture. The TCN decoder is comprised of stacked TCN layers \cite{Gehring_Auli_Grangier_Yarats_Dauphin_2017, bai2018empirical}, which directly operate on the temporal dimension of the output embedding from the encoder. The TCN decoder expands the output embedding as needed for prediction. Unlike recurrent units and attention mechanisms, the TCN decoder relies on convolution operations in the feature space, resulting in a smaller number of parameters. This makes the TCN decoder more parameter-efficient compared to recurrent units and attention mechanisms \cite{mohamed2020social}.

At the end of the decoder, we adopt Inverse Discrete Cosine Transformation (IDCT) \cite{Zhou_2011, Mao2019LearningTD} to generate the predicted motion displacement $Y_{{T}:T+N-1}$ for each individual and transform it back to pose space $X_{{T+1}:T+N}$. For ultra-long-term prediction, our method has the capability to directly utilize the predicted poses of the previous $N$ frames as input, allowing for autoregressive prediction of the subsequent frames.

\paragraph{Loss function.}
We first utilize a reconstruction loss $\mathcal{L}_{rec}$ based on the GJPE to constraint global pose dynamics. To prevent distorted poses, we then introduce a pose constraint loss $\mathcal{L}_{pose}$ to further optimize predicted poses. The above loss functions are formulated as 
\begin{equation}
\label{eq_loss}
\mathcal{L}_{rec} = \frac{1}{J*N}\sum_{i=T+1}^{T+N}{\sum_{j=1}^{J}{||\hat{y}_{i,j} - y_{i,j}||^2}}, \quad\quad
\mathcal{L}_{pose} = AJPE(X_{T+1:T+N}, \hat{X}_{T+1:T+N}),
\end{equation}
where $\hat{y}_{i,j}$ and $y_{i,j}$ are ground-truth and predicted displacement at time $i$. $J$ is the number of body joints. For one training sample, we jointly optimize by
\begin{equation}
\label{eq_loss_all}
\mathcal{L}_{joint} = \mathcal{L}_{rec} + \alpha \mathcal{L}_{pose},
\end{equation}
where $\alpha$ is a hyper-parameter.

\section{Additional Training and Implementation Details}
\label{app:implementation}
For all baselines, we utilize a 25-frame (1.0s) input sequence to generate predictions for both short-term and long-term scenarios, consisting of 25 frames (1.0s). For ultra-long-term prediction, we employ an autoregressive strategy to realize it. All of them are trained for 50 epochs with a batch size of 32 and a learning rate of 0.0003. All the baselines are implemented in PyTorch, and the experiments are performed on a Nvidia GeForce RTX 3090 GPU. We provide details below for the different hyperparameters used in each baseline.
\paragraph{HRI method.}
The \texttt{kernel\_size} of convolutional layers is set as 5 and \texttt{dct\_n} (width of DCT) is set as 10.  For GCN predictor, the \texttt{d\_model} (hidden dimension) and \texttt{num\_stage} (number of layers) are 256 and 12, respectively. The \texttt{output\_n} (output frames) is set as 5, which means the model requires 5 iterations for 25 frames of prediction. 
\paragraph{MRT method.}
The number of output frames is set as 25. The other hyperparameters of the model are based on the original paper and have not been changed in any way. Please refer to \cite{wang2021multi} for detail.
\paragraph{TBIFormer method.}
The \texttt{kernel\_size} of convolutional layers and the number of output frames are set as 5 and 25, respectively. The other hyperparameters of the model are based on the original paper. Please refer to \cite{peng2023trajectory} for detail.
\paragraph{SocialTGCN method.}
For the PSM module, the \texttt{kernel\_size} and \texttt{stride} of 1D convolutional layers are 10 and 1. There are 12 stack GCN layers for the PSM and 3 stacked SoicalTGCN encoder layers. The layer-number of the TCN deocder is 5. The hidden dimension of GCN in PSM and SocialTGCN are 265 and 512, respectively. The number of output frames is set as 25. The $\alpha$ for $\mathcal{L}_{pose}$ in joint loss is 0.1.

\section{Additional Quantitative Results}
\label{app:quantitative}
\begin{figure}[h]
\begin{minipage}[b]{0.5\linewidth} 
    \centering
    \includegraphics[width=1.0\linewidth]{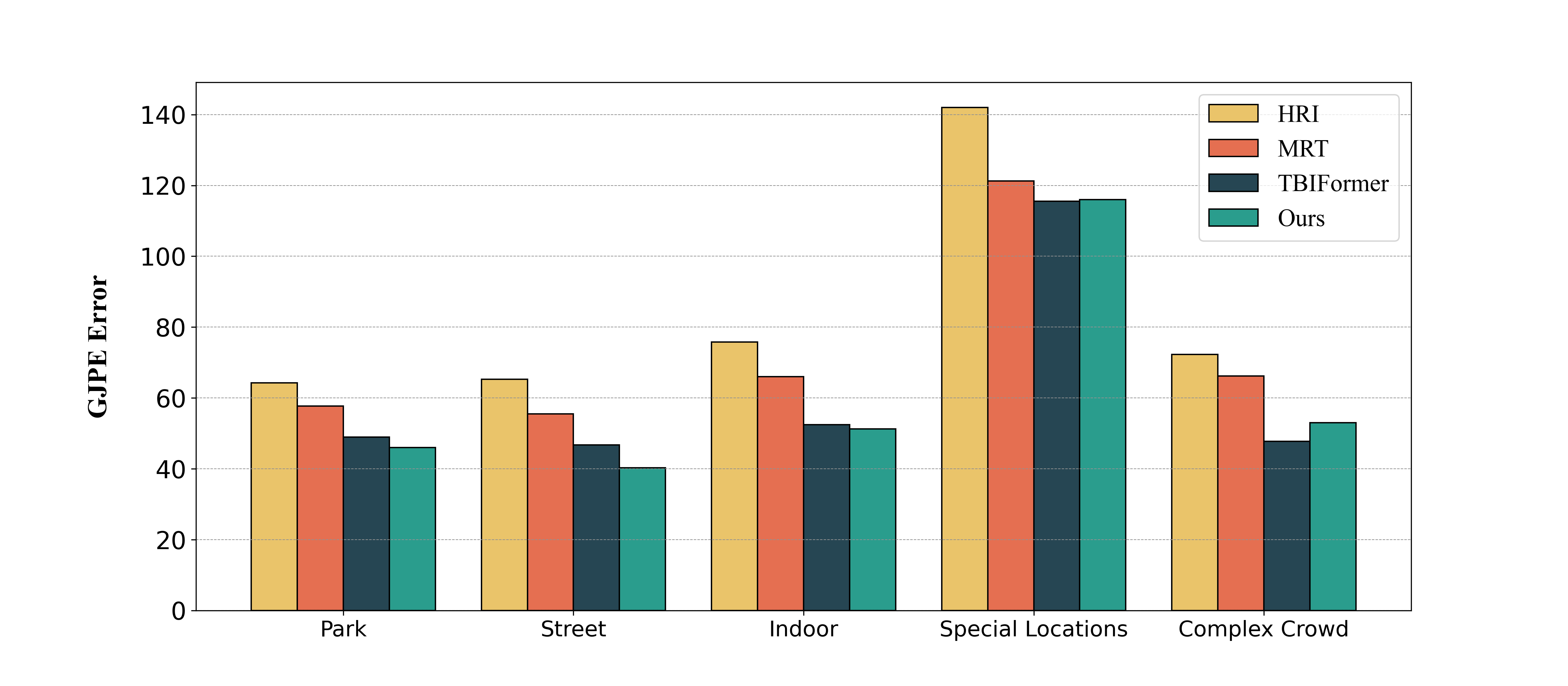}
  \end{minipage} 
  \begin{minipage}[b]{0.5\linewidth} 
    \centering
    \includegraphics[width=1.0\linewidth]{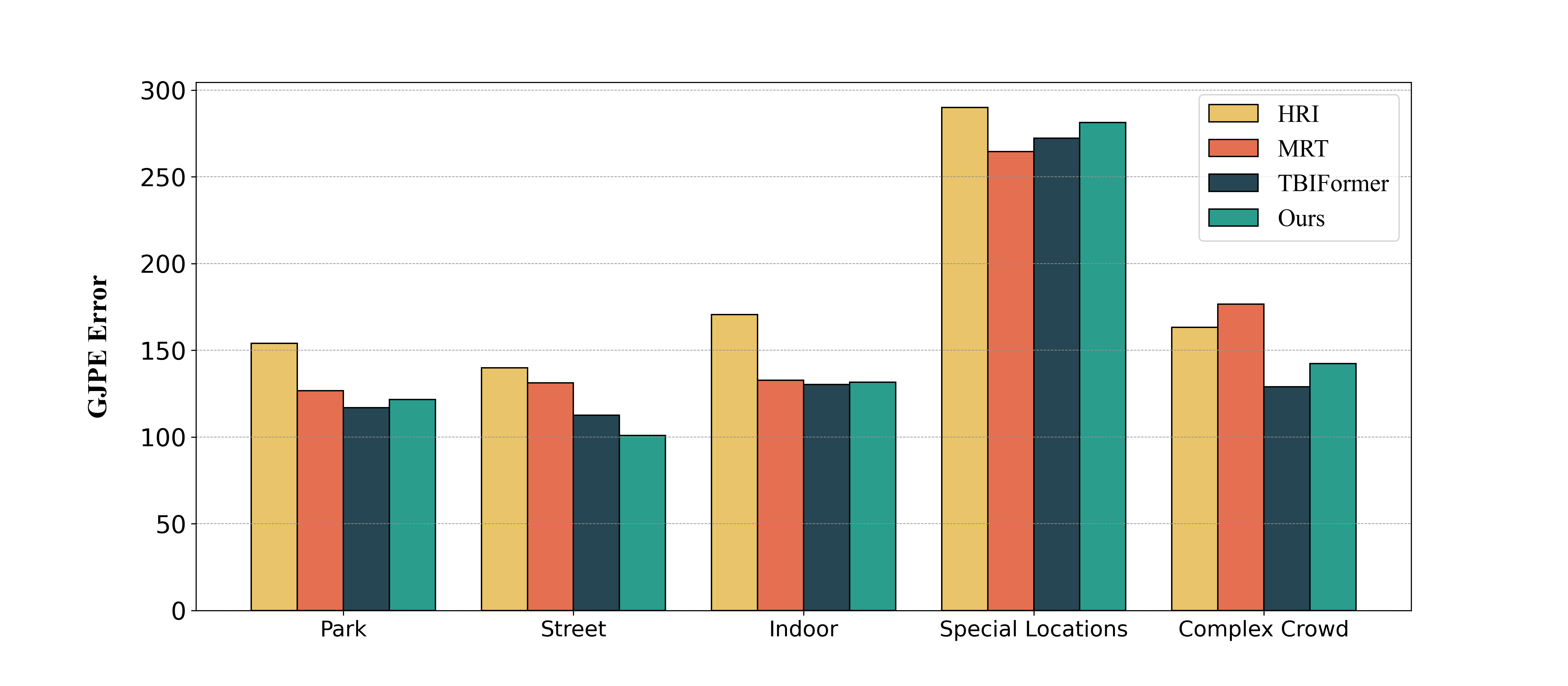}
  \end{minipage} 
\caption{The left figure is mean GJPE error for short-term prediction comparing with the baselines, while the right one is for long-term prediction.}
\label{fig:mean_GJPE} 
\end{figure}

\begin{figure}[h]
    \centering
    \includegraphics[width=.9\linewidth]{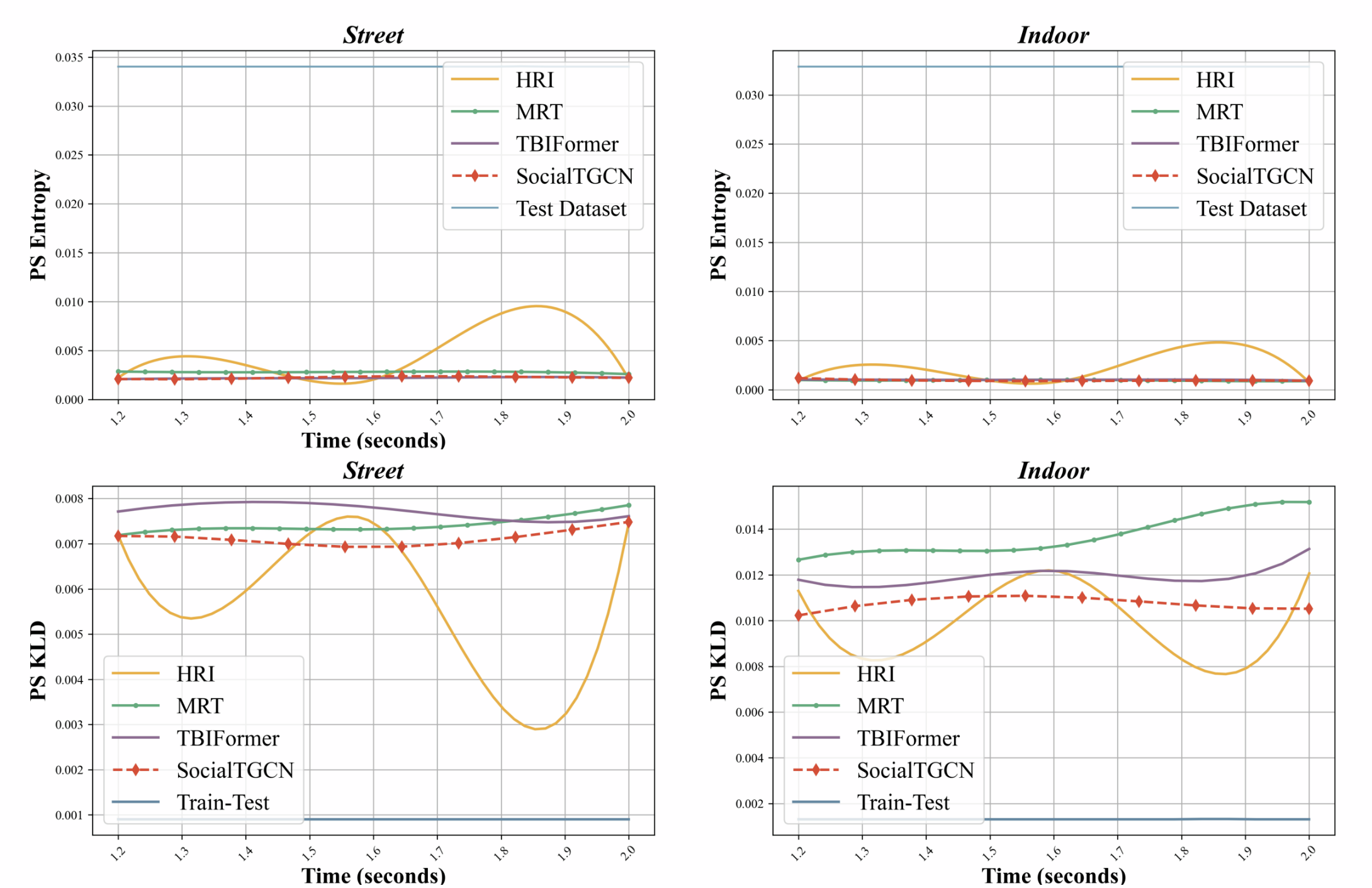}
    \caption{Power Spectrum (PS) results for the other two scenes, \textit{Street} and \textit{Indoor}.  }
    \label{fig:app-ps}
    \vspace{-5mm}
\end{figure}

Firstly, we provide the average GJPE errors for short-term and long-term prediction in Figure~\ref{fig:mean_GJPE}. As we can see, our method also get competitive results. The results of ultra-long-term prediction for the other scenes in PS metrics are shown in Figure~\ref{fig:app-ps}.

We further conduct extensive ablation studies on MI-Motion to investigate the contribution of key technical components in the proposed SocialTGCN, with results in Table~\ref{tab:ablat_short_term} and Table~\ref{tab:ablat_long_term}. The pose constraint loss is introduced to optimize generated poses. When it is removed, we can observe a significant performance decrease on AJPE results. The SocialGCN encoder plays a crucial role in our method, as it is the core component that captures the social interaction among individuals and incorporates it into the motion prediction process. Without the SocialGCN encoder, the performance of our method would be compromised, resulting in sub-optimal predictions.

\begin{table}[t] 
\begin{center}
  \caption{
  \textbf{Ablation results of the proposed SocialTGCN on short-term horizon.} The error is evaluated by the three metrics in millimeter. Column-wise best result in bold.}
  \label{tab:ablat_short_term}
  \resizebox{\linewidth}{!}
  {
  \begin{tabular} {cl aaaa cccc aaaa cccc aaaa}
    \multicolumn{2}{c}{}&\multicolumn{4}{c}{\cellcolor{SceneBlue}\textit{Park}} & \multicolumn{4}{c}{\cellcolor{SceneRed}\textit{Street}} & \multicolumn{4}{c}{\cellcolor{SceneOrange}\textit{Indoor}} & \multicolumn{4}{c}{\cellcolor{ScenePurple}\textit{Special Locations}} & \multicolumn{4}{c}{\cellcolor{SceneGreen}\textit{Complex Crowd}}\\ \toprule 
    \multicolumn{2}{c}{Time (ms)} & 80 & 160 & 320 & 400 & 80 & 160 & 320 & 400  & 80 & 160 & 320 & 400 & 80 & 160 & 320 & 400 & 80 & 160 & 320 & 400 \\  \midrule
    \multirow{3}{*}{\rotatebox{90}{\textbf{GJPE}}} &  W/o pose contraint loss & \multicolumn{1}{a}{\textbf{18}} & \textbf{34} & 61 & 73 & 18 & 33 & 63 & 78 & \textbf{19} & 37 & 67 & 81 & \textbf{45}& 90& \textbf{161}&\textbf{191}& 20& 39& 74& 90 \\ 
     &W/o SoicalGCN endoer &  \multicolumn{1}{a}{\textbf{18}} & \textbf{34} & \textbf{60} & \textbf{72} & \textbf{15} & \textbf{28} & 55 & 67 & \textbf{19} & \textbf{36} & \textbf{62} &\textbf{75}  &46 &90 &166 &199 &\textbf{19} &38 &72 &87\\ 
     &Full Method  &  \multicolumn{1}{a}{\textbf{18}} &\textbf{34} & \textbf{60} &\textbf{72}  & \textbf{15} & \textbf{28} & \textbf{54} & \textbf{64} & 20 & 37 & 67 &81 &\textbf{45} &\textbf{89} &165 &199 &20 &\textbf{37} &\textbf{70} &\textbf{85}\\ 
    \midrule
    \multirow{3}{*}{\rotatebox{90}{\textbf{AJPE}}} &  W/o pose contraint loss  & \multicolumn{1}{a}{16} & 31 & 54 & 62 & 14 & 26 & 50 & 60 & 18 & 34 & 61 & 72 & 42 & 79 & 130 & 146 & 16 & 32 & 60 & 70 \\ 
     &W/o SoicalGCN endoer  &  \multicolumn{1}{a}{\textbf{14}} & \textbf{26} & \textbf{44} & \textbf{51} & 12 &\textbf{20} &\textbf{38} &\textbf{46} & 16 & 28 &\textbf{48} &\textbf{57} &38 &72 &119 &136 &\textbf{14} &\textbf{26} &\textbf{47} &57  \\ 
     &Full Method &  \multicolumn{1}{a}{\textbf{14}} & \textbf{26} &45 &53 & \textbf{11} & 21 &\textbf{38} & \textbf{46} &\textbf{15} & \textbf{27} & 49 &58  &\textbf{36} &\textbf{69} &\textbf{117} &\textbf{134} &\textbf{14} &\textbf{26} &48 &\textbf{56}\\ 
     \midrule
    \multirow{3}{*}{\rotatebox{90}{\textbf{RFDE}}} &  W/o pose contraint loss & \multicolumn{1}{a}{\textbf{17}} & \textbf{31} & 57 & 67 & 17 & 30 & 56 & 69 & 20 & 40 &80 &97 & \textbf{39} & 75 & \textbf{134} & \textbf{160} & 19 & 37 & 70 & 86 \\ 
     &W/o SoicalGCN endoer  &  \multicolumn{1}{a}{18} &32 &55 &65 &14 & 24 & 47 & 56 &\textbf{19} & 38 & \textbf{70} &\textbf{87} &41 & 79 & 146 & 173  &\textbf{18} &\textbf{32} &63 &79 \\ 
     &Full Method &  \multicolumn{1}{a}{18} & \textbf32 & \textbf{53} & \textbf{64} & \textbf{13} & \textbf{23} &\textbf{41} &\textbf{50} &20 &\textbf{36} & 73 &90 &\textbf{39} &\textbf{74} &142 &174 &\textbf{18} &\textbf{32} &\textbf{62} &\textbf{77} \\ 
     \bottomrule
  \end{tabular}
  }
\end{center}

\begin{center}
  \caption{
  \textbf{Ablation results of the proposed SocialTGCN on long-term horizon.} The error is evaluated by the three metrics in millimeter. Column-wise best result in bold.}
  \label{tab:ablat_long_term}
  \resizebox{\linewidth}{!}
  {
  \begin{tabular}{cl aaa ccc aaa ccc aaa}
    \multicolumn{2}{c}{}&\multicolumn{3}{c}{\cellcolor{SceneBlue}\textit{Park}} & \multicolumn{3}{c}{\cellcolor{SceneRed}\textit{Street}} & \multicolumn{3}{c}{\cellcolor{SceneOrange}\textit{Indoor}} & \multicolumn{3}{c}{\cellcolor{ScenePurple}\textit{Special Locations}} & \multicolumn{3}{c}{\cellcolor{SceneGreen}\textit{Complex Crowd}}\\ \toprule 
    \multicolumn{2}{c}{Time (ms)} & 560 & 720 & 1000 & 560 & 720 &1000 & 560 & 720 & 1000 & 560 & 720 &1000 & 560 & 720 & 1000  \\   
    \midrule

    \multirow{3}{*}{\rotatebox{90}{\textbf{GJPE}}} &  W/o pose contraint loss & \textbf{93} & \textbf{111} & \textbf{141} & 104 & 121 & 144 & 103 & 122 &\textbf{141} & \textbf{232} &\textbf{259} & \textbf{296} & 118 & 144 & 185 \\ 
     &W/o SoicalGCN endoer &  {96} & 117 & 159 & 87 &102 &130 &\textbf{99} & \textbf{117} &147 & 246 &284  & 331 & 115 & 140 & \textbf{172}  \\ 
     &Full Method  &  95 &116 &154 & \textbf{81} & \textbf{98} & \textbf{124} & 108 & 127 &160 & 246 &276 & 322 & \textbf{113} & \textbf{137} &177  \\  
    \midrule
    \multirow{3}{*}{\rotatebox{90}{\textbf{AJPE}}} & W/o pose contraint loss & {76} & 87 & 103 &76 & 87 & 96 & 88 & 101 &115 & 163 & 169 & 181 &86 & 97 & 116  \\ 
     &W/o SoicalGCN endoer  & \textbf{65} & 77 & 97 & 59 &67 &78 & \textbf{71} &\textbf{82} &99 &160 & 174 & 190 &72 & 83 & 99   \\ 
     &Full Method  & 66 & \textbf{76} & \textbf{93} & \textbf{58} & \textbf{66} &\textbf{76} &73 &83 &\textbf{97} &\textbf{154} &\textbf{161} &\textbf{174} & \textbf{70} &\textbf{81} &\textbf{97}  \\ 
     \midrule
    \multirow{3}{*}{\rotatebox{90}{\textbf{RFDE}}} &  W/o pose contraint loss & {91} & 114 & \textbf{148} & 92 & 115 & 152 & 124 & 145 & 169 &\textbf{207} &\textbf{243} & \textbf{298} & 113 & 144 & 199   \\ 
    
     &W/o SoicalGCN endoer  & 89 &115 &163 & 73 &91 &124 &\textbf{113} & \textbf{131} &\textbf{168} &219 &270 &337 & 108 & 135 &178   \\ 
    
     &Full Method  &  {\textbf{88}} &\textbf{112} &154 &\textbf{62} &\textbf{79} &\textbf{108} & 119 & 138 &180 & 226 & 265 & 321 & \textbf{106} &\textbf{131} & \textbf{173}  \\ 
     \bottomrule 
  \end{tabular}
  }
\end{center}
\vspace{-5mm}
\end{table}

\section{Additional Qualitative Results}
\label{app:qualitative}
\begin{figure}[t]
  \centering
  \includegraphics[width=.87\linewidth]{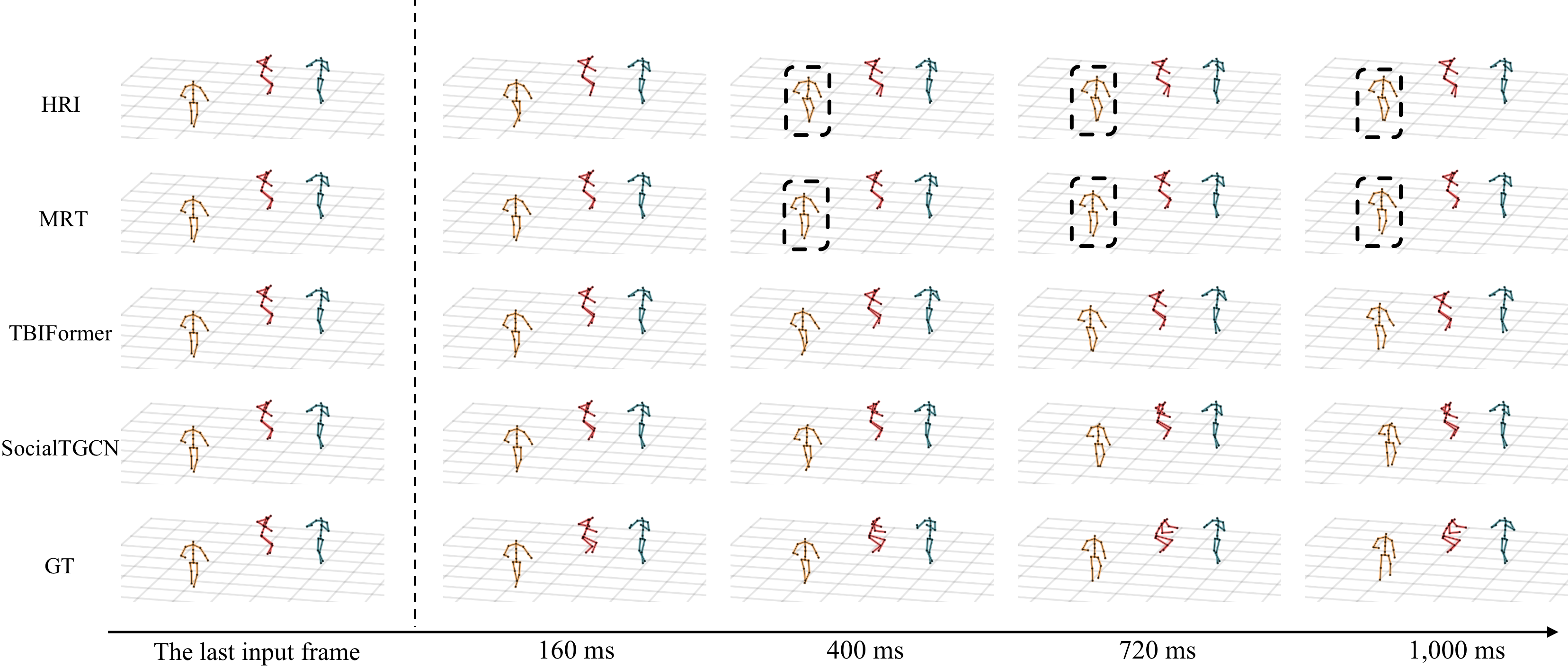}
  \caption{
  \textbf{Qualitative visualization for the \textit{Street} scene.} The left one column are inputs, and the right four columns are predictions.
  }
  \label{fig:app_quality_vis1}
\end{figure}

\begin{figure}[h]
  \centering
  \includegraphics[width=.87\linewidth]{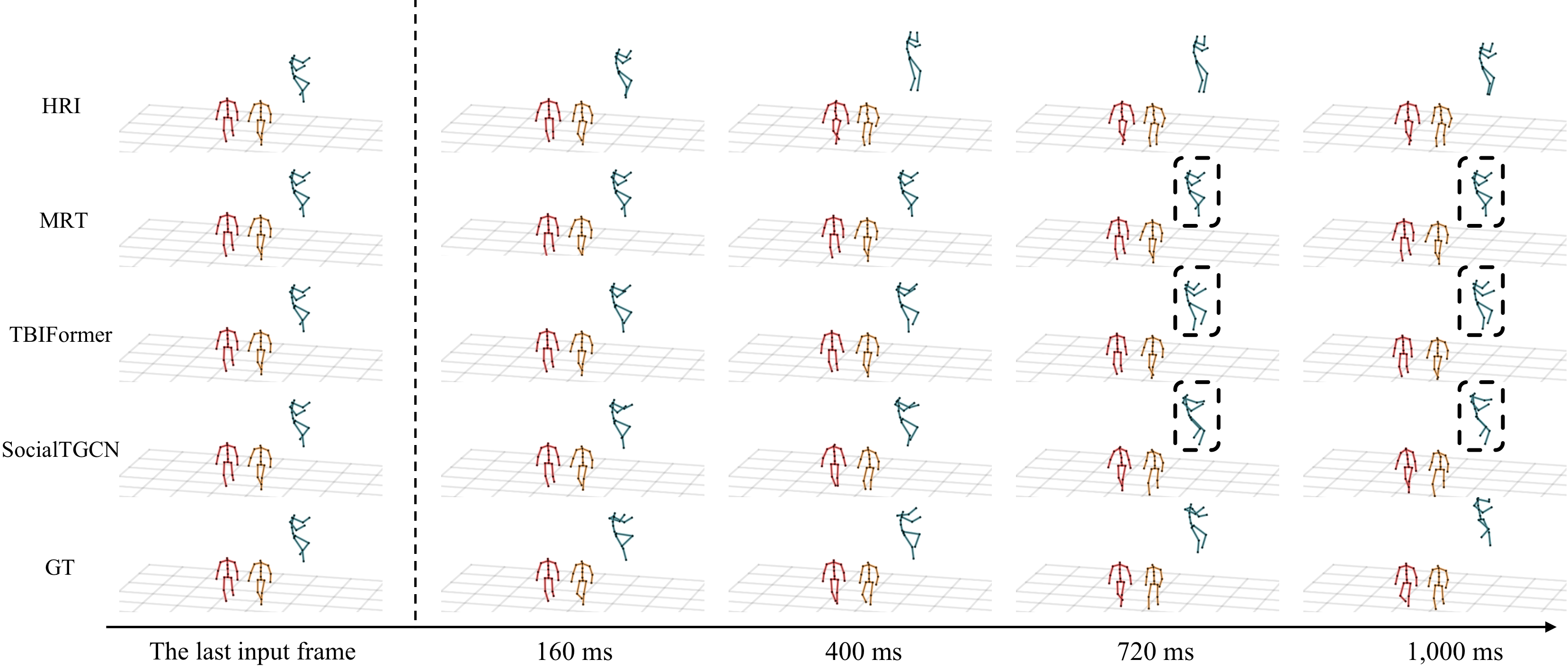}
  \caption{
  \textbf{Qualitative visualization for the \textit{Indoor} scene.} The left one column are inputs, and the right four columns are predictions.
  }
  \label{fig:app_quality_vis2}
\end{figure}

\begin{figure}[h]
  \centering
  \includegraphics[width=.87\linewidth]{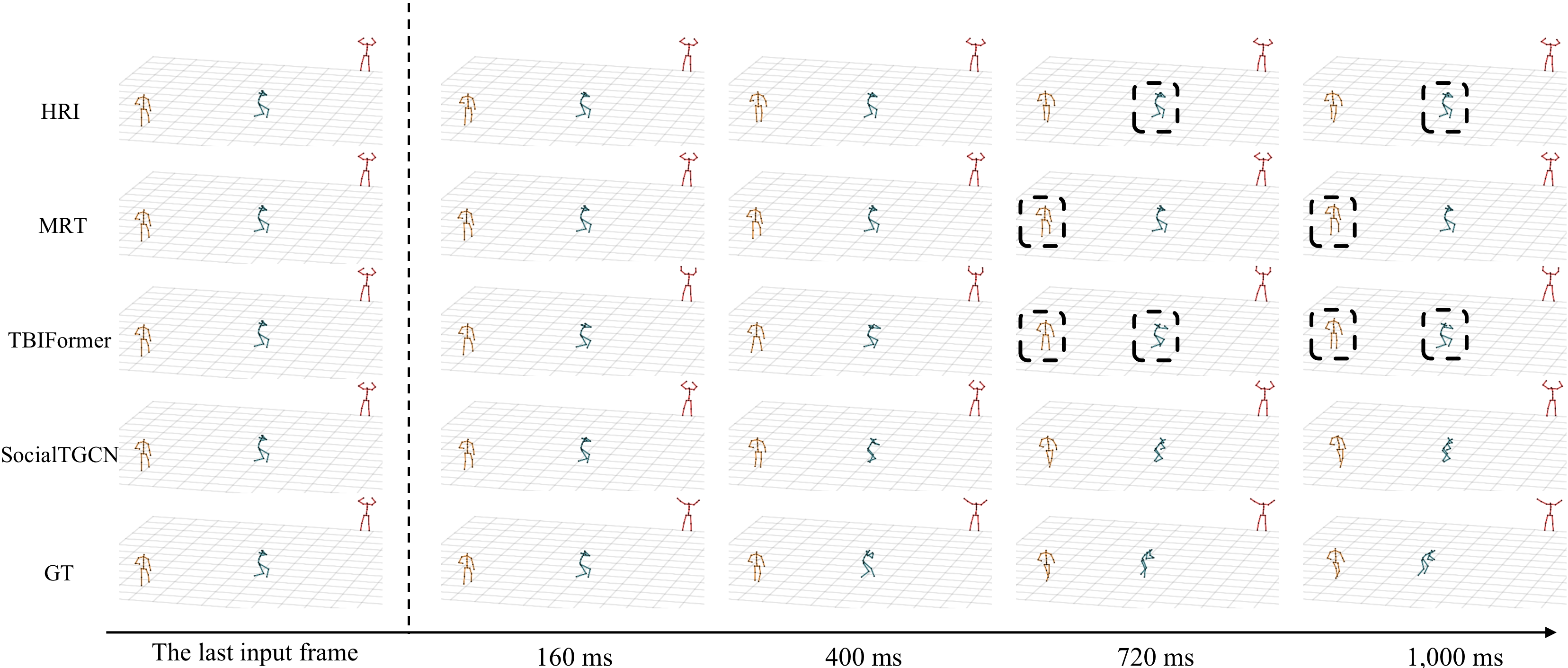}
  \caption{
  \textbf{Qualitative visualization for the \textit{Special Locations} scene.} The left one column are inputs, and the right four columns are predictions.
  }
  \label{fig:app_quality_vis3}
\end{figure}

\begin{figure}[h]
  \centering
  \includegraphics[width=0.8\linewidth]{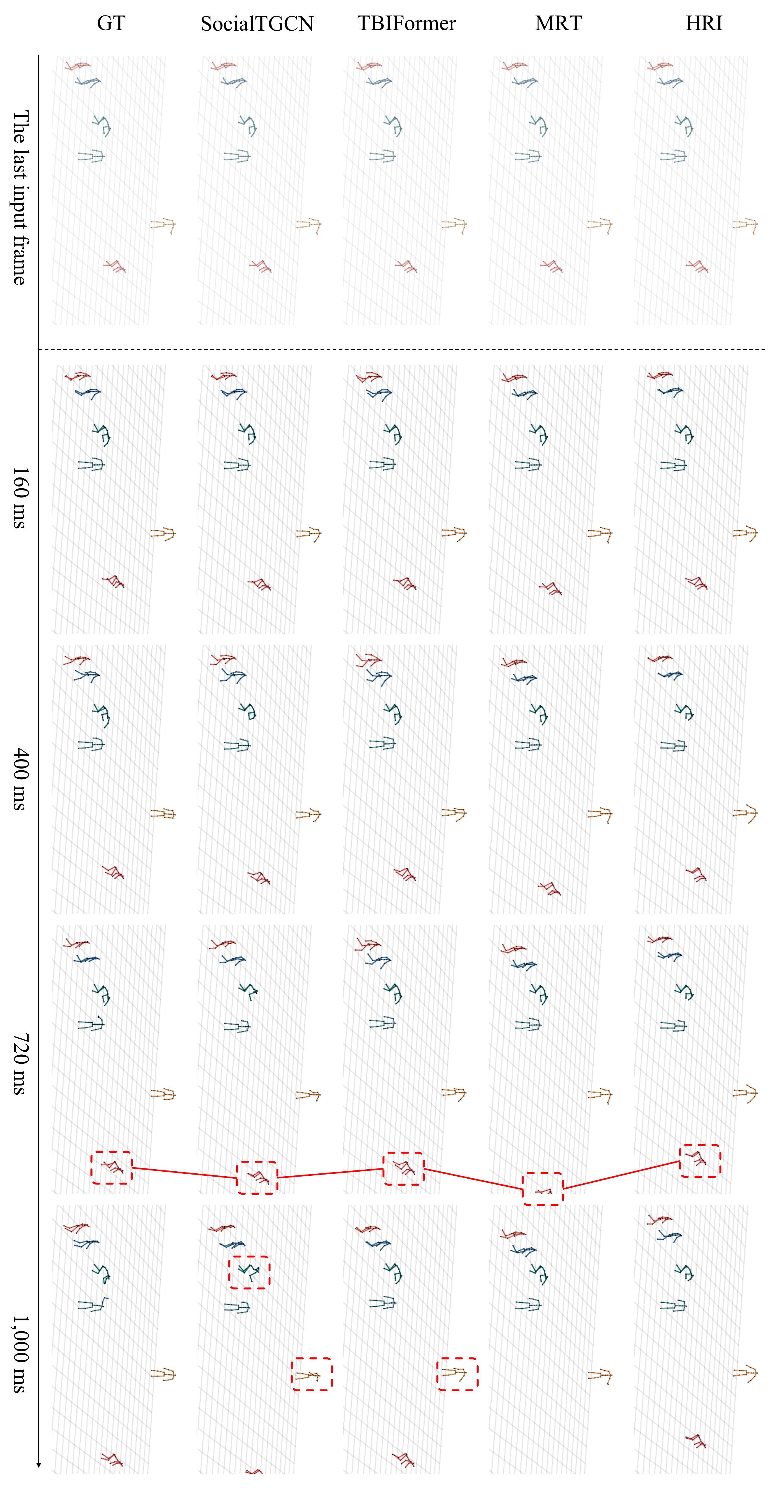}
  \caption{
  \textbf{Qualitative visualization for the \textit{Crowd} scene.} The left one column are inputs, and the right four columns are predictions.
  }
  \label{fig:app_quality_vis4}
\end{figure}

 We supplement the qualitative results of the other scenes, as shown in Figure~(\ref{fig:app_quality_vis1}, \ref{fig:app_quality_vis2}, \ref{fig:app_quality_vis3} and \ref{fig:app_quality_vis4}). We compare our model with other baselines, \ie HRI\cite{mao2020history}, MRT\cite{wang2021multi} and TBIFormer\cite{peng2023trajectory}. Our experimental results demonstrate that our method consistently produces smoother and more natural motion predictions compared to other approaches. Additionally, our predictions are found to be closer to the ground truth, indicating the effectiveness of our method in capturing the underlying dynamics of human motion. For further qualitative analysis and evaluation, we provide additional results for ultra-long-term scenarios on the dataset website.

\section{Dataset Accessibility and Long-Term Preservation Plan} \label{app:dataset-accessibility}
\vspace{-2mm}
The instructions to access our dataset is summarized on our project page at \href{https://mi-motion.github.io/}{\texttt{https:/mi-motion.github.io/}}.

The dataset is stored on the Google Drive and Baidu disk, which can provide secure and convenient access to your files. These platforms can provide stable accessibility and long-term preservation of our dataset.
\vspace{-2mm}
\section{Author Statement of Responsibility} \label{app:author-statement}
\vspace{-2mm}
The authors confirm all responsibility in case of violation of rights and confirm the licence associated with the data and code.

\newpage
\vspace{-10mm}

\end{document}